\pdfoutput=1
\documentclass[11pt]{article}
\usepackage{setspace}
\usepackage[]{acl}

\usepackage{times}
\usepackage{latexsym}

\usepackage[T1]{fontenc}
\usepackage[utf8]{inputenc}

\usepackage{microtype}

\usepackage{inconsolata}

\usepackage{booktabs}
\usepackage{amsmath}
\usepackage{xcolor}
\usepackage{graphicx}
\usepackage{subcaption}
\usepackage[cjk]{kotex}
\usepackage{wasysym}
\usepackage{makecell}
\usepackage{multirow} 
\usepackage{mdframed}
\usepackage{tablefootnote}
\usepackage{longtable}

\mdfsetup{
linecolor=white,
backgroundcolor=gray!20,
font=\small
}

\newcommand\crehate{CREHate}

\title{Exploring Cross-Cultural Differences in English Hate Speech Annotations:\\From Dataset Construction to Analysis}
\author{Nayeon Lee$^1$, Chani Jung$^{1,*}$, Junho Myung$^{1,}$\Thanks{ Equal contribution.}, Jiho Jin$^1$, \\\textbf{Jose Camacho-Collados$^2$, Juho Kim$^1$, Alice Oh$^1$} \\
  $^1$KAIST, $^2$Cardiff University \\
  \texttt{\{nlee0212, 1016chani, junho00211, jinjh0123\}@kaist.ac.kr,} \\ \texttt{camachocolladosj@cardiff.ac.uk, juhokim@kaist.ac.kr,  alice.oh@kaist.edu}}

\begin{document}
\setlength\belowcaptionskip{-1.2ex}
\captionsetup{skip=4.9pt}
\begin{spacing}{0.98}

\maketitle
\begin{abstract}
\textit{\textbf{Warning}: this paper contains content that may be offensive or upsetting.}
% Juho's comments

Most hate speech datasets neglect the cultural diversity within a single language, resulting in a critical shortcoming in hate speech detection. 
% and other culturally sensitive tasks.
To address this, we introduce \textbf{\crehate{}}, a \textbf{CR}oss-cultural \textbf{E}nglish \textbf{Hate} speech dataset.
To construct \crehate{}, we follow a two-step procedure: 1)~cultural post collection and 2)~cross-cultural annotation.
We sample posts from the SBIC dataset, which predominantly represents North America, and collect posts from four geographically diverse English-speaking countries (Australia, United Kingdom, Singapore, and South Africa) using culturally hateful keywords we retrieve from our survey.
Annotations are collected from the four countries plus the United States to establish representative labels for each country.
Our analysis highlights statistically significant disparities across countries in hate speech annotations.
Only 56.2\% of the posts in \crehate{} achieve consensus among all countries, with the highest pairwise label difference rate of 26\%.
Qualitative analysis shows that label disagreement occurs mostly due to different interpretations of sarcasm and the personal bias of annotators on divisive topics.
% label disagreements tend to come from the ambiguity and the subjectivity of the posts.
Lastly, we evaluate large language models (LLMs) under a zero-shot setting and show that current LLMs tend to show higher accuracies on Anglosphere country labels in \crehate{}.
% \todo{Lastly, we develop cross-cultural hate speech classifiers that are more accurate at predicting each country's labels than the monocultural classifiers. This confirms the utility of \crehate{} for constructing culturally sensitive hate speech classifiers.}
Our dataset and codes are available at: \url{https://github.com/nlee0212/CREHate}
\end{abstract}

\section{Introduction}
\begin{figure}[t!]
    \centering
    \includegraphics[width=0.85\columnwidth]{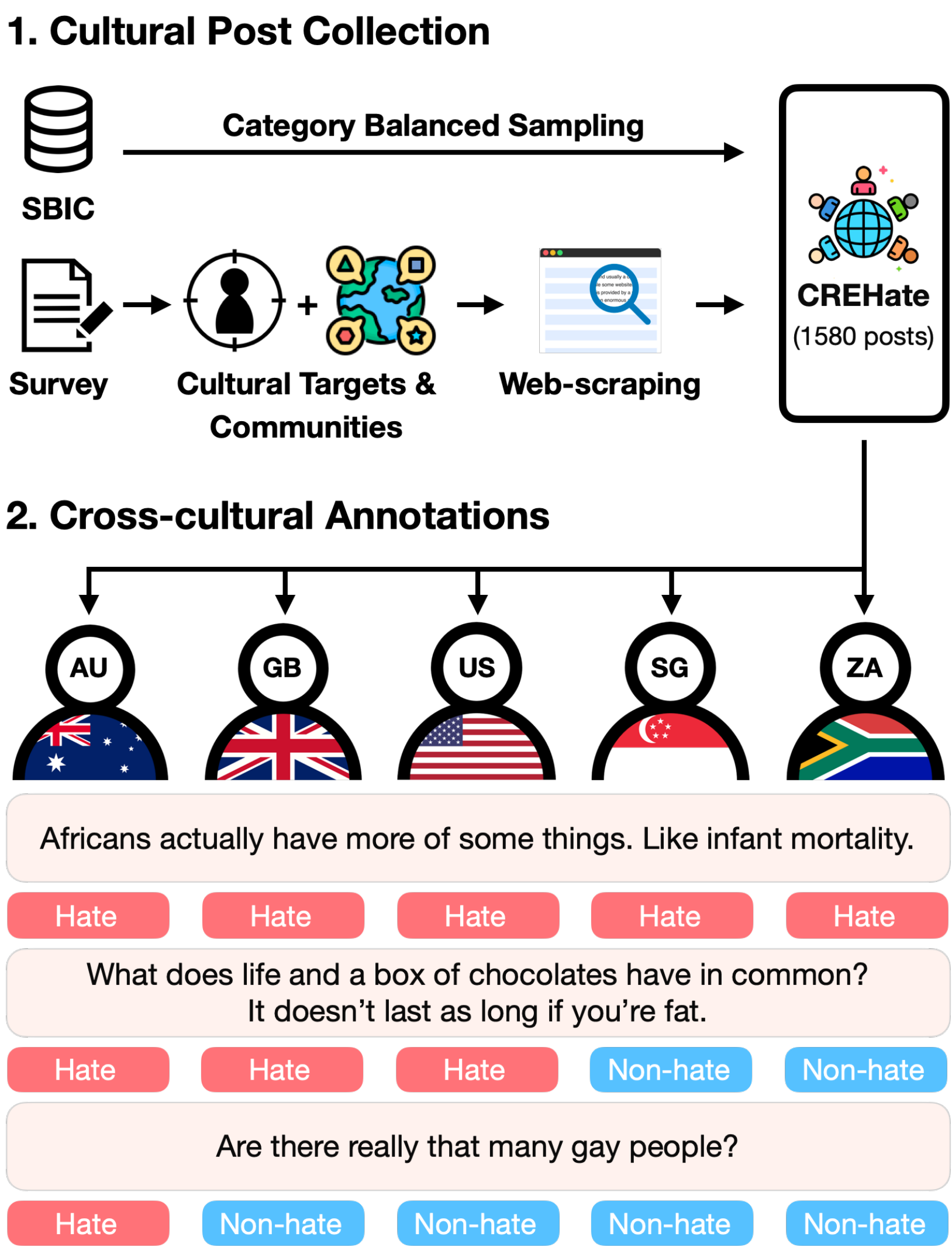}
    \caption{Illustration of the two-step procedure of \crehate{} construction: 1) cultural post collection and 2) cross-cultural annotation. The examples show how annotations on identical posts differ across countries.}
    \label{fig:annotation}
\end{figure}
Identifying hate speech is highly subjective and relies heavily on an annotator's understanding and knowledge of the cultural context~\cite{10.1145/3308560.3317083,waseem-2016-racist-tacl}.
Unfortunately, existing English hate speech datasets often overlook the cultural diversity within the posts and the annotators.
They are predominantly collected from Twitter (Table \ref{tab:1_hatespeech}), reflecting a disproportionate representation of certain countries, notably the United States\thinspace\footnote{The US has the most Twitter users by country (\url{https://datareportal.com/essential-twitter-stats}).}.
Furthermore, annotators' geographic location is either neglected or limited to only one or two countries, despite English being spoken in over 50 countries\thinspace\footnote{The World Factbook, Languages (\url{https://www.cia.gov/the-world-factbook/field/languages/})}. 
This limitation hinders the datasets' ability to capture diverse viewpoints.
Figure \ref{fig:annotation} illustrates how people from different countries show varying hate speech annotations on identical posts.
% \add{Therefore, we aim to explore the impact of cultural diversity on hate speech based on a dataset reflecting the diversity.}

% \add{
% Looking at five countries does not necessarily capture the full extent of the cultural diversity of English speakers, and there are cultural differences even within the same country.
% However, throughout this paper, we align culture with countries and assume that annotators from specific countries understand and represent the main cultural norms. Thus, their geographic diversity leads directly to cultural diversity. 
% We deliberately select five countries from vastly different regions of the globe to show how an individual's cultural background impacts their interpretations, specifically within the domain of hate speech.
% We aim to provide a more representative understanding of hate speech in diverse cultural contexts.

Our research aims to investigate the influence of cultural diversity on hate speech. To achieve this, we construct a dataset that reflects diversity and examine how cultural background affects the interpretation of hate speech by annotators. Specifically, we align culture with nationality when exploring how cultural background influences annotators' interpretations of hate speech. We acknowledge that focusing only on cross-country differences may not fully encompass the multifaceted cultural dynamics within each country. However, it offers a starting point to understand how annotators' cultural background based on nationality affects language interpretation, particularly in sensitive areas like hate speech. 
This approach underlines the importance of further, more detailed studies into the complex interplay of cultural identities and their impact on language perception~\cite{Kramsch2014}, especially for enhancing hate speech moderation on global platforms.

% This paper analyzes hate speech annotation differences across five countries, chosen from vastly different parts of the world, to cover cultural diversity in English speakers.
% We align culture with nationality to explore how cultural background influences interpretations of hate speech, although we acknowledge that this may not encompass the multifaceted cultural dynamics within each country.
% Since language is closely linked to cultural identity~\cite{Kramsch2014}, we examine how this affects the interpretation of hate speech. 
% Although our analysis may not capture the full extent of the cultural diversity among English speakers globally, it highlights the crucial role of cultural background in interpreting language. 
% This underscores the importance of including diverse cultural perspectives in datasets, particularly in the sensitive domain of hate speech.
% }
% More details about the annotators are in Section \S\ref{sec:3_dataset_construction}.

To this end, we construct \textbf{\crehate{}}---a \textbf{CR}oss-cultural \textbf{E}nglish \textbf{Hate} speech dataset---comprising 1,580 online posts annotated by individuals from five English-speaking countries: Australia (AU), United Kingdom (GB), Singapore (SG), the United States (US), and South Africa (ZA)\thinspace\footnote{Two-letter ISO country codes (\url{https://www.iso.org/iso-3166-country-codes.html}).}.
Construction of \crehate{} is done in a 2-step procedure: 1) cultural post collection and 2) cross-cultural annotation (Figure \ref{fig:annotation}). 
For cultural post collection, we collect 600 posts from YouTube and Reddit using keywords gathered from surveys from four countries: AU, GB, SG, and ZA. 
We also sample 980 posts from SBIC~\cite{sap-etal-2020-social}, a toxic language dataset of social media posts including diverse target groups, primarily reflecting a North American perspective (Table \ref{tab:1_hatespeech})\thinspace\footnote{Reddit and Gab's users are mainly from the US (\url{https://www.semrush.com/website/reddit.com/overview/}, \url{https://www.semrush.com/website/gab.com/overview/}), as well as Twitter.}.
For cross-cultural annotation, five annotators from each country annotate each post to establish representative labels for each country. 
Based on cross-cultural considerations, this dataset creation procedure makes \crehate{} more culturally comprehensive than datasets that ignore cultural differences within English-speaking countries.
% The labels are used to analyze cross-cultural differences in hate speech annotation.
% The overview of the dataset construction process is shown in .

\begin{table}[t!]
\centering
% \small
\resizebox{\columnwidth}{!}{
\begin{tabular}{@{}cccc@{}}
\toprule
\textbf{Datasets}                                                                                          & \textbf{\begin{tabular}[c]{@{}c@{}}Post\\ Source\end{tabular}}                                                                & \textbf{\begin{tabular}[c]{@{}c@{}}Source\\ Country\end{tabular}} & \textbf{\begin{tabular}[c]{@{}c@{}}Annotation\\Platform\\ (Country)\end{tabular}} \\ \midrule
\begin{tabular}[c]{@{}c@{}}MLMA\\ \cite{ousidhoum-etal-2019-multilingual}\end{tabular}                     & Twitter                                                                        & US*                                                                     & {\begin{tabular}[c]{@{}c@{}}MTurk\\ (N/A)\end{tabular}}                                                                      \\ \midrule
\begin{tabular}[c]{@{}c@{}}ImplicitHateCorpus\\ \cite{elsherief-etal-2021-latent}\end{tabular}             & Twitter                                                                        & US                                                                     & {\begin{tabular}[c]{@{}c@{}}MTurk\\ (N/A)\end{tabular}}                                                                      \\ \midrule
\begin{tabular}[c]{@{}c@{}}SBIC\\ \cite{sap-etal-2020-social}\end{tabular}                                 & \begin{tabular}[c]{@{}c@{}}Twitter,\\ Reddit, Gab,\\ Stormfront\end{tabular} & US*                                                                     & {\begin{tabular}[c]{@{}c@{}}MTurk\\ (US, CA\footnote{})\end{tabular}}                                                                   \\ \midrule
\begin{tabular}[c]{@{}c@{}}HateXplain\\ \cite{Mathew_Saha_Yimam_Biemann_Goyal_Mukherjee_2021}\end{tabular} & \begin{tabular}[c]{@{}c@{}}Twitter,\\ Gab\end{tabular}                         & US*                                                               & {\begin{tabular}[c]{@{}c@{}}CrowdFlower\\ (N/A)\end{tabular}}                                                                      \\ \midrule
\begin{tabular}[c]{@{}c@{}}OLID\\ \cite{zampieri-etal-2019-predicting}\end{tabular}                        & Twitter                                                                        & US*                                                               & {\begin{tabular}[c]{@{}c@{}}CrowdFlower\\ (N/A)\end{tabular}}                                                                      \\ \midrule
\citet{Davidson_Warmsley_Macy_Weber_2017}                                                                  & Twitter                                                                        & US*                                                              & {\begin{tabular}[c]{@{}c@{}}CrowdFlower\\ (N/A)\end{tabular}}                                                                      \\ \midrule
\citet{founta2018large}                                                                                    & Twitter                                                                        & US*                                                               & {\begin{tabular}[c]{@{}c@{}}CrowdFlower\\ (N/A)\end{tabular}}                                                                      \\ \midrule
\textbf{CREHate (Ours)}                                                                                             & \begin{tabular}[c]{@{}c@{}}Twitter,\\ Reddit, Gab,\\ Stormfront,\\ YouTube\end{tabular}              & \begin{tabular}[c]{@{}c@{}}AU, GB,\\ SG, US*, ZA \end{tabular}       & \begin{tabular}[c]{@{}c@{}}MTurk,\\ Prolific, Tictag\\(AU, GB,\\ SG, US, ZA)\end{tabular}           \\ \bottomrule
\end{tabular}
}
\caption{Datasets for toxic language detection annotated using crowdsourcing platforms. Existing datasets neglect or limit the cultural backgrounds of the annotators and posts. `US*' means there is a high possibility that the post sources are biased towards US due to the platform's skewed user demographics, even if not explicitly targeted during the data collection stage.}
\label{tab:1_hatespeech}
\end{table}
\footnotetext{CA refers to Canada. }

We show that cross-cultural annotations of \crehate{} demonstrate significant differences across countries.
Only 56.2\% of the entire posts receive unanimous label agreement across all five countries, and the average pairwise agreement between countries is 78.8\%, with a maximum label disagreement of 26.0\%. 
The pairwise label agreement distribution among countries exhibits a notable deviation from that of randomly selected annotator groups, with its average being 2.58$\sigma$ lower than the average pairwise label agreement of the random groups.
Furthermore, by conducting a qualitative analysis of potential reasons for label disagreements, we show that the primary contributing factors are likely due to different understandings of sarcasm and the personal bias of annotators on divisive topics.

Finally, we show that current LLMs tend to show higher accuracy scores on core Anglosphere country labels in \crehate{}.
We further identify the limitations of these models in culture-specific hate speech classification, in which they are asked to predict hate speech based on the target country.

% \todo{Additionally, we present hate speech classifiers that can be tailored to different cultural contexts with the help of \crehate{}.
% These classifiers employ various training techniques, including multi-labeling, multi-task learning, and culture tagging, to discern distinct labels for each country within a unified model.
% Our approach achieves an accuracy improvement of up to 8.2\% when compared to the separate models trained on each country's labels, presenting a step towards creating more equitable and culturally sensitive automated content moderation systems.}

Our main contributions are as follows:
\begin{itemize}
    
    \item We build \crehate{}, a cross-cultural English hate speech dataset including posts and annotations from diverse cultural backgrounds.

    \item Through quantitative and qualitative analysis, we identify significant variations in hate speech annotations attributed to the cultural backgrounds of the posts and the annotators.
    
    \item We show LLMs' higher accuracies on core Anglosphere country labels in hate speech classification and limitations in making culture-specific predictions.
    % \item \todo{We adopt various techniques to construct culturally adaptive hate speech classifiers using \crehate{}}.

\end{itemize}

\section{Related Work}

\noindent \textbf{Impact of Annotator Demographics. } 
Annotator demographics, such as gender, affect their annotations in NLP datasets \cite{biester-etal-2022-analyzing}.
Hate speech detection is particularly a subjective task where the demographics can affect the annotations, inter-annotator agreement (IAA), and classifier performance \cite{waseem-2016-racist-tacl,sap-etal-2022-annotators,goyal2022your,larimore-etal-2021-reconsidering, binns2017like}.

\noindent \textbf{Cultural Considerations in Hate Speech Detection.}
\label{sec:related_cultural}
Recent research in offensive language examined cultural differences and built datasets in diverse languages \citep{lee-etal-2023-hate,jeong-etal-2022-kold,jin2023kobbq,arango-monnar-etal-2022-resources,deng-etal-2022-cold,demus-etal-2022-comprehensive,mubarak2022emojis}, but these papers assume that a single language reflects a single culture. 
However, languages such as English are spoken by a culturally diverse population, necessitating the consideration of cultural differences among language speakers.
\citet{arango-monnar-etal-2022-resources} built the first hate speech dataset for Chilean Spanish to enrich the cultural diversity of Spanish datasets.
They evaluated knowledge transfer performance on another Spanish dataset with a different cultural background, but the impact of cultural background on annotations was unexplored.
We aim to conduct a thorough study of how hate speech and its annotations vary across English-speaking countries. 

\noindent \textbf{Multiple Cultures in English NLP. }
\citet{frenda-etal-2023-epic} developed a corpus for irony detection, focusing on which annotator demographic group's perspectives are more represented by majority voting.
They collected posts and gathered annotators from five English-speaking countries: Ireland, India, AU, GB, and US.
Our study, focusing on hate speech detection, extends the scope by collecting posts as well as annotations from different cultures and investigating the annotation disparities stemming from cultural variations.

\section{Dataset Construction}

\label{sec:3_dataset_construction}
\begin{table}[t!]
\centering
\small
% \resizebox{\columnwidth}{!}{
\begin{tabular}{@{}lllc@{}}
\toprule
\multicolumn{2}{c}{\textbf{Data}}                   & \textbf{Source}   & \textbf{\# Posts} \\ \midrule
\multirow{8}{*}{CREHate} & \multirow{5}{*}{CC-SBIC} & Reddit            & 568               \\
                         &                          & Twitter           & 273               \\
                         &                          & Gab               & 80                \\
                         &                          & Stormfront        & 59                \\ \cmidrule(l){3-4} 
                         &                          & \textit{subtotal} & \textit{980}      \\ \cmidrule(l){2-4} 
                         & \multirow{3}{*}{CP}      & Reddit           & 311               \\
                         &                          & YouTube           & 289               \\ \cmidrule(l){3-4} 
                         &                          & \textit{subtotal} & \textit{600}      \\ \midrule
                         &                          & \textbf{total}    & \textbf{1,580}    \\ \bottomrule
\end{tabular}
% }
\caption{Data statistics and sources of \crehate{}. CC-SBIC refers to cross-culturally re-annotated SBIC posts. CP refers to additionally collected cultural posts from four countries (AU, GB, SG, and ZA), which are also cross-culturally annotated.}
\label{tab:3_1_stats}
\end{table}
This section describes the construction process of \crehate{}, an English hate speech dataset with both posts and annotations collected from five different countries to analyze the country-level divergences when it comes to hate speech.
We follow a 2-step procedure:  1) cultural post collection and 2) cross-cultural annotation. 
The dataset consists of 1,580 posts, each with five labels representing five countries, resulting in a total of 7,900 labels.
Dataset statistics are shown in Table \ref{tab:3_1_stats}.

\begin{table*}[t!]
\centering
\resizebox{\textwidth}{!}{%
\begin{tabular}{l|l|c|c|c|c|c}
\toprule
\textbf{Data} & \textbf{Post} & \textbf{AU} & \textbf{GB} & \textbf{SG} & \textbf{US} & \textbf{ZA} \\ \midrule
CC-SBIC & \begin{tabular}[t]{@{}l@{}}Why did they plant trees in harlem? For free public transit. \end{tabular}& $\Circle$ & $\Circle$ & $\Circle$ & $\Circle$ & $\times$ \\ \specialrule{0.01em}{0.2em}{0.2em}
CP$_{AU}$ & \begin{tabular}[t]{@{}l@{}}The special treatment is for real abbos only....... equality. \end{tabular}& $\Circle$ & $\Circle$ & $\times$ & $\times$ & $\Circle$ \\ \specialrule{0.01em}{0.2em}{0.2em}
CP$_{GB}$ & \begin{tabular}[t]{@{}l@{}}@USER A pikey calling someone uneducated. The irony is painful. \end{tabular}& $\times$ & $\Circle$ & $\Circle$ & $\Circle$ & $\Circle$ \\ \specialrule{0.01em}{0.2em}{0.2em}
CP$_{SG}$ & \begin{tabular}[t]{@{}l@{}}The attitude of Malays being "holier than thou" is kinda sickening. Have experienced\\it many times, sadly it’s not gonna stop. \end{tabular}& $\Circle$ & $\Circle$ & $\times$ & $\times$ & $\Circle$ \\ \specialrule{0.01em}{0.2em}{0.2em}
CP$_{ZA}$ & \begin{tabular}[t]{@{}l@{}}I wonder when people will recognize the EFF for what it really is. It’s not a political\\party, it’s a terrorist organization.\end{tabular}& $\times$ & $\times$ & $\times$ & $\times$ & $\Circle$ \\ \bottomrule
\end{tabular}}
\caption{Examples of annotated posts in CREHate. $\Circle$ refers to `hate', and $\times$ refers to `non-hate' label. The example posts require culture-specific knowledge on target minorities (abbos, Malays), locations (Harlem), public organizations (EFF), and slang (pikey). People from different countries provide different hate speech annotations for posts including culture-specific context.}
\label{tab:examples}
\end{table*}
\noindent \textbf{English-speaking Countries. }
We choose one country from each continent to ensure geographical diversity while also considering cultural differences within and outside the Anglo-American sphere of influence~\cite{10.1002/polq.13068,4TheAngloAmericanWorldView1}. 
Specifically, we select three core Anglosphere countries---AU, GB, and US~\cite{6a1005f3-9282-3341-8afa-a9dbe0bd3b9a}---and two countries with English as official language but not necessarily the primary language---SG and ZA~\cite{KHOKHLOVA2015983,doi:10.1080/03050069728587}.

\subsection{CREHate Post Collection}

\subsubsection{Sampling from SBIC}
To incorporate hate speech targeting diverse groups, we sample posts from the SBIC dataset~\cite{sap-etal-2020-social}, which contains annotations of offensive posts targeted towards different demographic groups and minorities.
From SBIC, we sample 980 posts while balancing the target group categories.
The details of SBIC and the sampling process are specified in Appendix \ref{sec:A_1_1_sbic}.
This set of sampled posts is referred to as CC-SBIC (\textbf{C}ross-\textbf{C}ultural SBIC) throughout the paper, as it is cross-culturally re-annotated as mentioned in \S\ref{sec:analysis}.

\subsubsection{Collecting Cultural Samples}
\label{sec:3_1_2_Collecting_Cultural_Samples}
The sources of SBIC's posts are culturally skewed towards the US, resulting in a bias towards prevalent target groups and the cultural context of the US.
To address this issue, we collect and annotate 150 cultural online posts each (a total of 600 posts) from four English-speaking countries: AU, GB, SG, and ZA.
The posts are collectively referred to as CP, and the country-specific posts are called CP$_{AU}$, CP$_{GB}$, CP$_{SG}$, and CP$_{ZA}$, respectively.
% To address this potential cultural bias, we have collected 600 cultural online posts from four English-speaking countries: AU, GB, SG, and ZA.
% We select 150 cultural posts from each country, collectively referred to as CP and individually labeled as CP$_{AU}$, CP$_{GB}$, CP$_{SG}$, and CP$_{ZA}$. 

\noindent \textbf{Keyword Collection. }
To efficiently gather hate speech posts, we use words that refer to specific demographic groups that are often subjected to hate as queries.
We recruit workers whose nationality and current residency match our target country and who have spent most of their lives in their respective countries to obtain the most appropriate and culturally relevant keywords.
We ask them to provide commonly targeted groups and possible hateful keywords that may refer to them within their culture.
We collect target groups in \textit{race/ethnicity}, \textit{gender/sexuality}, and \textit{religion/culture} categories, the three main categories within the original SBIC dataset. 
% Limiting the number of categories allows us to collect sufficient posts from each category and culture.
We continue collecting until we gather at least 20 keywords per country.

\noindent \textbf{Post Collection. }
We gather popular social media and news sites from the workers in their countries and select Reddit as our primary social media platform for collecting comments, as it is widely used across all countries.
We also crawl comments from the YouTube channels of news sites in each country.
% \noindent \textbf{Post Filtering}
To ensure that we have enough potentially hateful posts in our dataset, we go through a pre-annotation stage, gathering only two annotations from the country the post originated from. 
Based on the pre-annotation results, we finalize 150 posts to be annotated from each country, maintaining the ratio of posts labeled as hate between 39.8\% and 48.5\% for each country\thinspace\footnote{Specific post crawling and sampling process is provided in Appendix \ref{sec:A_1_1_cultural_posts}.}.
As a result, the posts from each culture contain some unique topics and keywords, such as `\textit{abo}' or `\textit{lebs}' in CP$_{AU}$, `\textit{gypsy}' or `\textit{paki}' in CP$_{GB}$, `\textit{malay}' or `\textit{pinoy}' in CP$_{SG}$, and `\textit{boer}' or `\textit{EFF}' in CP$_{ZA}$.

% \noindent \textbf{Posts with Cultural Context}
% Through unique word extraction from posts from each of the four cultures, 

\subsection{Cross-Cultural Annotation}

\label{sec:analysis}

\begin{figure*}[t!]
\centering
\subfloat[]{%
 \includegraphics[width=0.8\columnwidth]{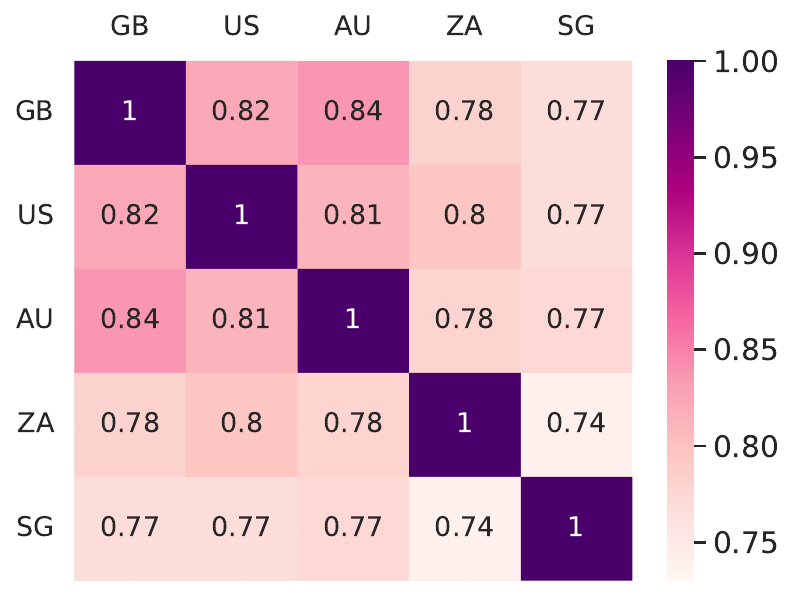}
  \label{fig:4_3_jp}
} 
% \hspace{1cm} %\qquad 
\subfloat[]{%
 \includegraphics[width=0.73\columnwidth]{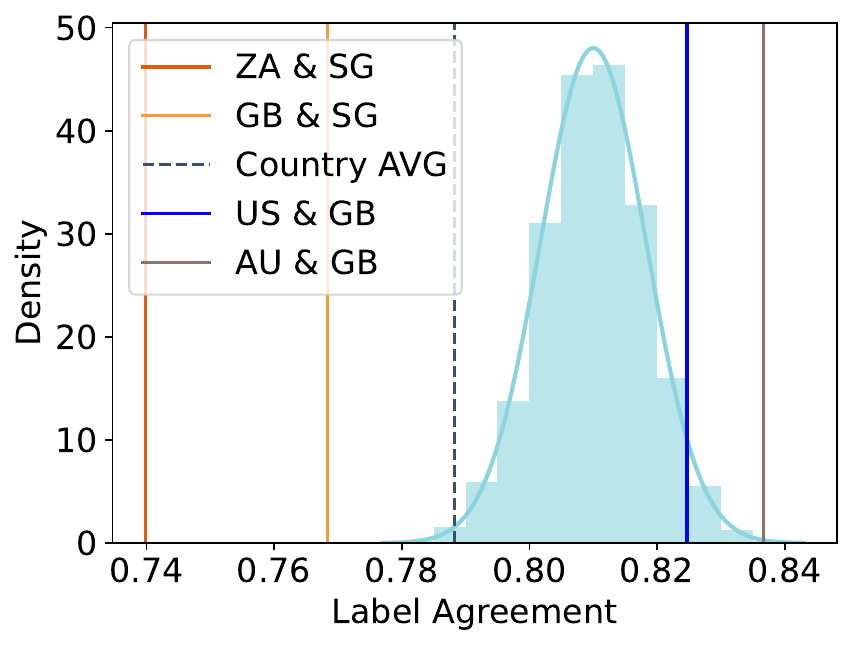}
 \label{fig:4_3_random}
}
\caption{(a) Pairwise label agreements across countries ordered by the average agreement with others. Labels from Singapore tend to be the most different. (b) Comparison of the label agreements among country pairs and random ones. The histogram and its density function show the distribution of pairwise label agreements among randomly selected annotator groups. The solid lines indicate country pairs with top-2 and bottom-2 label agreement scores, and the dashed line indicates the average of label agreements of all country pairs. Countries that are closely related exhibit high label agreements compared to the random annotator groups, whereas culturally distant countries show significantly low label agreements compared to label agreements from random annotator groups.}
\end{figure*}

\noindent \textbf{Annotator Recruitment. }
We recruit annotators from five countries, applying the same annotator qualifications as we used for keyword collection, from Prolific\thinspace\footnote{\url{https://www.prolific.co/}} (AU, GB, ZA), Amazon Mechanical Turk\thinspace\footnote{\url{https://www.mturk.com/}} (US), and Tictag\thinspace\footnote{\url{https://www.tictagkr.com/}} (SG) depending on annotator recruitment availability of the desired country.
As a result, we have 1,061 annotators, balancing their gender but not restricting others for a broader representation of demographics~\cite{frenda-etal-2023-epic}.
Table \ref{tab:a_annotator_demographic} shows a detailed demographic distribution of annotators.

\noindent \textbf{Annotation Process. }
Before annotating, annotators are required to review the definitions\thinspace\footnote{\url{https://www.un.org/en/hate-speech/understanding-hate-speech/what-is-hate-speech}} and examples of hate and non-hate speech. 
Examples are selected among posts with identical labels across all countries from the pilot study.
The task is to annotate posts as either \textit{Hate} or \textit{Non-hate}, with an additional option of \textit{I don't know}\thinspace\footnote{The \textit{I don't know} labels took up about 3-7\% of the raw annotations, and more analysis on these labels are mentioned in Appendix \ref{sec:A_2_idk_analysis}.}. 
We obtain five \textit{Hate} or \textit{Non-hate} labels for each post from each country.
The specific annotation process and quality control methods are in Appendix \ref{sec:quality_control}.
% Attention-check questions are incorporated throughout the annotation tasks to ensure high-quality data collection.
% To prevent a single annotator from having an excessive effect on the dataset, we limit the number of annotations from each annotator to less than 5\% of the total annotations.

\noindent \textbf{Label Finalization. }
After gathering all five annotations, we use majority voting to finalize the representative labels for each country.
% Our analysis shows that there exists a moderate agreement between each annotator and the aggregated labels with an average Cohen's $\kappa$ agreement of 0.640~\cite{kappa_article}.
% In addition, we include soft labels in the final dataset for future research purposes, although we did not incorporate them into our current work.
Examples of posts with labels from each country are presented in Table \ref{tab:examples}.

\section{Analysis on the Annotations}
In this section, we show that varying cultural backgrounds of annotators and posts lead to a significant disparity in hate speech annotation.

\subsection{Significance of Cultural Backgrounds}
\label{sec:4_1_statistical_analysis}
%To show that an annotator's cultural background is a significant factor in hate speech detection, 
To analyze the role of an annotator's cultural background in hate speech detection, we obtain labels representative of different demographic categories\footnote{For more details on the demographic categories analyzed and their statistics, please refer to Table \ref{tab:a_annotator_demographic} in the Appendix.} %mentioned in Table \ref{tab:a_annotator_demographic}
using majority voting.
We only collect labels from groups with at least three annotators per post on average.
Labels from each group are subjected to chi-squared tests, and the results indicate significant disparities in annotations across country ($p = 0.000$), race ($p = 0.002$), gender ($p = 0.006$), and education level ($p = 0.000$), while there were no significant differences for other groups.
Several studies have shown the importance of race or gender of annotators~\cite{pei-jurgens-2023-annotator,10.1145/3531146.3533216}, whereas the impact of annotators' cultural background has been underexplored.

\subsection{Label Agreement among Countries}
\label{sec:4_2_label_agreement_country}
% \noindent \textbf{Posts with Varying Agreements}
% We assess the level of agreement among different countries regarding the proportion of posts that received each level of agreement.
% Only 56.2\% of the posts achieved unanimous agreement among all countries, with 25.5\% of the posts receiving agreement among four countries.

\noindent \textbf{Pairwise Country Label Agreement. }
Overall, only 56.2\% of the posts achieve unanimous agreement across all countries, with 25.5\% of the posts showing agreement across four countries.
To further explore the label differences across cultures, we examine the label agreements between all pairs of countries, as shown in Figure \ref{fig:4_3_jp}.
% Figure \ref{fig:4_3_jp} shows the label agreement between all pairs of countries.
It suggests pairwise label agreements among core Anglosphere countries are greater than those observed in other country pairs.
Among all countries, AU and GB exhibit the highest label agreement at 83.7\%, while SG and ZA show the lowest agreement at 74.0\%.

We compare these results to the cultural distance index~\cite {kogut1988effect}\thinspace\footnote{A value of 0 indicates identical cultural norms, while a value close to 1 indicates average distance among all countries.} between countries, which measures the degree to which cultural norms in two countries differ (Table \ref{tab:c_cultural_distance_index}). 
The cultural distance and the hate speech label agreements among the countries show a high negative Pearson correlation with \(r=-0.658\) ($p=0.039$).
This implies that country pairs with more considerable cultural distances have lower label agreement.
SG and ZA, the country pair with the lowest label agreement, show a higher cultural distance (2.178) than AU and GB (0.144), the country pair with the highest agreement.

Furthermore, to investigate the pairwise label differences on identical posts across different countries, we employ the McNemar Test~\cite{McNemar1947}. 
The results indicate significant pairwise label disparity between 8 out of 10 country pairs.
% Further analysis using the McNemar Test on the two divisions of \crehate{} is discussed in Appendix \ref{sec:c_pairwise_label_analysis}.
% Furthermore, our examination reveals more pronounced differences within CP than in SBIC posts. 
% Specifically, CP$_{AU}$, CP$_{GB}$, and CP$_{ZA}$ exhibit significant differences in eight or more country pairs out of ten.
% CP$_{SG}$ shows significant differences in six out of ten country pairs. 
% However, in SBIC posts, significant differences are observed in only two pairs: US and ZA, and AU and SG.
% These outcomes demonstrate significant variations in annotations among individuals from diverse countries, particularly within culturally specific posts.

\noindent \textbf{Comparison with Random Annotator Groups. }
To show that label disparities stem from the annotators' cultural backgrounds rather than random variations among individuals, we compare the pairwise country label agreements with the distribution of label agreements between randomly organized annotator groups.
For each post, we create two groups of five randomly selected annotations out of 25 (5 from each country) and construct representative labels from each group via majority voting.
We calculate the label agreement of the two groups for the whole dataset and repeat this process $10^5$ times.
% The outcomes are illustrated in Figure \ref{fig:4_3_random}, which includes a histogram and the corresponding estimated normal distribution.
The outcomes of this comparison, illustrated in Figure \ref{fig:4_3_random}, include a histogram and an estimated normal distribution curve of the label agreements among these random groups. 
Based on the D'Agostino-Pearson normality test \cite{272b2fa8-f3b3-371d-b34e-a4c6938458a1}, the label agreements among random annotators follow a normal distribution with $\mu=0.81$ and $\sigma=0.008$.

% The average label agreement between pairs of different countries is 0.79, falling 2.58$\sigma$ below the average agreement of the random annotator groups.
% This clearly illustrates a notable distinction between the two distributions.
% The two highest and two lowest label agreements between country pairs are marked in solid vertical lines. 
% The two highest pairwise country label agreements, shown between the core Anglosphere countries, are larger than the average label agreement among random annotator groups by 1.77$\sigma$ and 3.22$\sigma$.
% The two lowest agreements fall 5.01$\sigma$ and 8.44$\sigma$ below this average. 
% The average label agreement between pairs of different countries falls 2.62$\sigma$ below the average agreement of the random annotator groups.
% Through this, we demonstrate that label agreements between culturally distant countries significantly differ from those between random annotator groups.
A critical observation from our analysis is the notable disparity in label agreements when comparing different countries. Specifically, the two highest label agreements, observed between core Anglosphere countries (US \& GB, AU \& GB), exceed the average agreement of the random groups by 1.77$\sigma$ and 3.22$\sigma$, respectively. However, the two lowest agreements observed between culturally distant countries (GB \& SG, ZA \& SG) fall significantly below this average, at 5.01$\sigma$ and 8.44$\sigma$. This pronounced disparity, along with the average pairwise label agreement between different countries being 2.62$\sigma$ lower than the average for random groups, strongly indicates that variations in perceptions of hate speech are not merely random differences among individuals. Instead, they are significantly influenced by cultural factors, showing more consistency within Anglosphere countries but substantial variation among other countries. This underscores the critical role of cultural contexts in hate speech detection and annotation across different English-speaking countries.
\begin{table}[t]
\centering
\small
% \resizebox{\columnwidth}{!}{
\begin{tabular}{l|ccc}
\toprule
            & \textbf{Agreement} & \textbf{H-F1} & \textbf{N-F1}  \\ \midrule
\textbf{CREHate} & 0.7882 & 0.7636 & 0.8077 \\ \midrule
\hspace{3mm}\textbf{CC-SBIC}    & 0.8045 & 0.8034 & 0.8050 \\
\hspace{3mm}\textbf{CP}     & 0.7617 & 0.6762 & 0.8108 \\ \midrule
\hspace{3mm}\textbf{CP$_{AU}$}      & 0.7293 & 0.6937 & 0.7565 \\
\hspace{3mm}\textbf{CP$_{GB}$}      & 0.7493 & 0.6851 & 0.7913 \\
\hspace{3mm}\textbf{CP$_{SG}$}      & 0.7827 & 0.6583 & 0.8390 \\
\hspace{3mm}\textbf{CP$_{ZA}$}      & 0.7853 & 0.6565 & 0.8433 \\ \bottomrule 
\end{tabular}
% }
\caption{The average pairwise label agreement scores (Agreement), F1 scores for hate (H-F1), and non-hate (N-F1) labels among all country pairs. 
 Our cultural posts (CP) show lower average pairwise country label agreement and lower F1 scores for hate labels. 
}
\label{tab:4_jp_f1}
\end{table}

\noindent \textbf{Label Agreements on Subsets of CREHate. }
We also analyze label agreements among countries on different subsets of CREHate (Table \ref{tab:4_jp_f1}).
Firstly, we compare the label agreements on two disjoint subsets of CREHate, CC-SBIC and CP.
Our findings reveal that CP has a lower average pairwise label agreement than CC-SBIC.
Although the two divisions show similar average pairwise F1 scores for non-hate labels, the F1 score for hate labels on CP significantly lags behind CC-SBIC's. 
This implies that CP derives more considerable label disparities for identifying one post as hate compared to CC-SBIC.
This trend is consistent across all sets of posts collected from different countries.

\noindent \textbf{Annotator Agreement. }
% IAA within a country also differs among countries.
Krippendorf's $\alpha$ is used to calculate IAA in US ($\alpha=0.462$), GB ($\alpha=0.425$), AU ($\alpha=0.408$), ZA ($\alpha=0.351$), and SG ($\alpha=0.344$).
These are higher than or comparable to those achieved in previous work in toxic language detection \citep{ross2016hatespeech,sap-etal-2020-social}.

\begin{table}[t!]
\centering
% \renewcommand{\arraystretch}{1.1}
% \small
\resizebox{\columnwidth}{!}{
\begin{tabular}{@{}l|l@{}}
\toprule
\textbf{Categories}              & \textbf{Subtypes}                                                                                                            \\ \midrule
Sloppy Annotation                & noise                                                                                                                        \\ \midrule
Ambiguity                        & \begin{tabular}[c]{@{}l@{}}analogy, false assertion, rhetorical question,\\ sarcasm, word play, reported speech\end{tabular} \\ \midrule
Missing Information              & ungrammatical, no context, not complete                                                                                      \\ \midrule
Subjectivity                     & personal bias, swearing, threatening                                                                                         \\ \bottomrule
% \textbf{Cultural Context} & requires cultural background                                                                                                 \\ 
\end{tabular}
}
\caption{Taxonomy of annotators’ disagreement in subjective tasks. We annotate the possible reasons behind label disagreements between countries, on top of culture-relevance labeling.}
\label{tab:4_3_disagreement_types}
\end{table}
\begin{table}[t!]
\centering
\resizebox{\columnwidth}{!}{%

\begin{tabular}{@{}c|l|c@{}}
\toprule
\textbf{Data} & \textbf{Post}                                                                                                                                                                                                      & \textbf{\begin{tabular}[c]{@{}c@{}}Reason\\ Type\end{tabular}} \\ \midrule
CP$_{AU}$       & \begin{tabular}[c]{@{}l@{}}You’re Irish and you support your pals\\ the Provisional IRA murderers of the\\ innocent Australian civilians Stephen\\ Melrose and Nick Spanos. You support\\ terrorists.\end{tabular}          & \begin{tabular}[c]{@{}c@{}}personal\\ bias\end{tabular}        \\ \midrule
CP$_{GB}$       & Gays have cards? are they sparkly?                                                                                                                                                                                           & sarcasm                                                        \\ \midrule
CP$_{SG}$       & \begin{tabular}[c]{@{}l@{}}So with the repeal of 377A, married men\\ are going to church to have gay sex?\end{tabular}                                                                                                     & \begin{tabular}[c]{@{}c@{}}not\\ complete\end{tabular}         \\ \bottomrule
\end{tabular}
}
\caption{Examples of disagreement reason annotation. For a sampled set of posts that countries disagree on, we annotate the possible reasons behind the disagreements following the disagreement reason taxonomy for subjective tasks by \citet{sandri-etal-2023}.}
\label{tab:4_3_disagreement_examples}
\end{table}
\subsection{Annotators' Disagreement Analysis}
\label{sec:4_3_disagreement_analysis}
We analyze the main factors behind label disagreements across countries using the taxonomy of reasons for annotators' disagreement for subjective tasks proposed by \citet{sandri-etal-2023}.
The categories and subtypes of the taxonomy are shown in Table~\ref{tab:4_3_disagreement_types}.
Appendix \ref{sec:A_D_disagreement_reason} shows detailed definitions and examples for each reason type.
Some of the annotated examples are shown in Table \ref{tab:4_3_disagreement_examples}.

% \begin{figure*}[t!]
% \centering
% % \textwidth
% \subfloat[]{%
%  \includegraphics[width=0.7\columnwidth]{Figures/ratio_of_culturespecific_posts.pdf}
%   \label{fig:4_3_ratio_cp}
% } 
% % \vspace{0.1cm} %\qquad 
% \subfloat[]{%
%  \includegraphics[width=0.7\columnwidth]{Figures/ratio_of_disagreement_cp.pdf}
%  \label{fig:4_3_ratio_dis_cp}
% }
% % \subfloat[]{%
% %  \includegraphics[width=0.65\columnwidth]{Figures/ratio_of_disagreement_noncp.pdf}
%  \label{fig:4_3_ratio_dis_noncp}
% % }
% \caption{(a) Ratio of culture-specific posts for each dataset division. Additionally collected cultural posts contain a higher ratio of culture-specific posts than posts from SBIC. (b), (c) shows the ratio of disagreement reasons within posts with and without culture-specific context, respectively. Sarcasm and personal bias contribute to the main factors of disagreement in both divisions.}
% \end{figure*}

\begin{figure}[t!]
    \centering
    \includegraphics[width=\columnwidth]{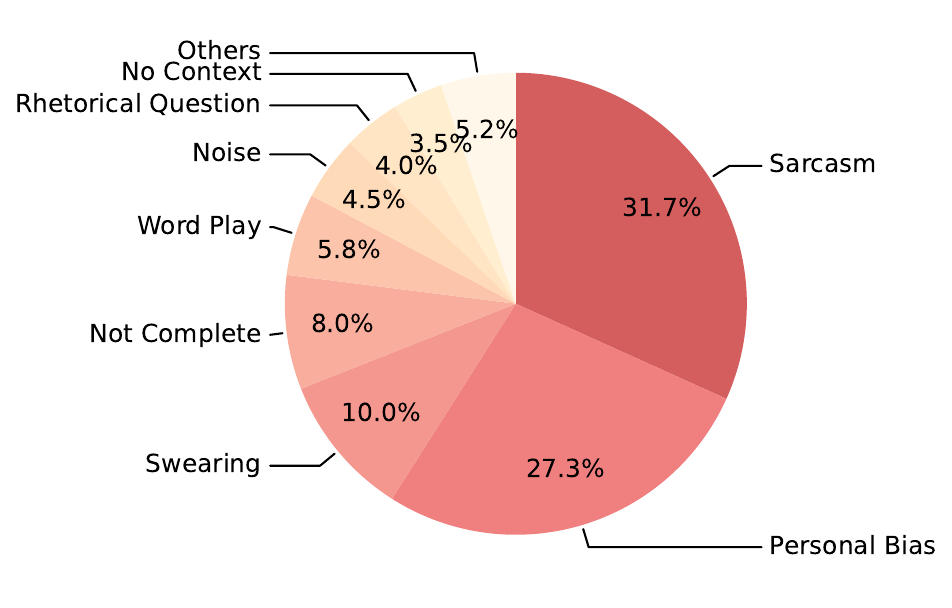}
    \caption{Ratio of disagreement reasons within posts. Differing interpretations of sarcasm and personal bias on divisive topics contribute to the main factors of disagreement.}
    \label{fig:4_3_ratio_dis}
\end{figure} 
\noindent \textbf{Disagreement Reason Annotation. }
Among the 1,580 posts in CREHate, 692 posts exhibit label discrepancies across countries.
To conduct a thorough analysis, we randomly sample 400 posts, including 200 posts from CC-SBIC and 50 posts from each of the four country's CP posts.
After a norming session, in which we clarify category definitions and apply them to our task, two authors annotate all sampled posts.
The initial Cohen's Kappa score from the two authors is 0.556, which is comparable to that of the annotations in \citet{sandri-etal-2023} (0.591), done by two linguists.
After that, the authors go through a discussion stage to establish a consensus on all labels.
As a result, the labels on the reasons for disagreement are finalized based on a unanimous agreement between the authors.

% \noindent \textbf{Impact of Culture-specific Context within Posts}
% Posts containing culture-specific context refer to culture-specific stereotypes about minorities, social phenomena, historical events, public figures, locations, or slang.
% Among the sampled posts of CREHate displaying label disagreements among countries, 47.0\% contained culture-specific context.
% Notably, 32.5\% of SBIC dataset posts and 61.5\% of additional posts included culture-specific context.
% Specifically, 72\% of the posts from ZA and 68\% from SG required cultural knowledge, compared to 48\% of posts from the GB and 58\% from AU.

\noindent \textbf{Possible Factors behind Disagreement. }
% In this section, we focus on the possible reasons behind the label disagreements for the posts with and without culture-specific context.
Overall, \textit{ambiguity} and \textit{subjectivity} of the posts contributed the most to the disagreements, taking up 44.3\% and 37.5\%, respectively.
% Disagreements within the posts were more likely to occur due to the \textit{ambiguity} of the posts (54.7\%) compared to \textit{subjectivity} (35.4\%). 
% In contrast, in the posts with culture-specific context, \textit{subjectivity} of the posts (41.5\%) was more likely to provoke disagreements than \textit{ambiguity} (39.4\%).
Among the lower-level subtype reasons, \textit{sarcasm} was the most frequently observed, followed by \textit{personal bias}, \textit{swearing}, and \textit{not complete} as shown in Figure \ref{fig:4_3_ratio_dis}.
A detailed analysis comparing CC-SBIC and CP's main disagreement reasons are shown in Appendix \ref{sec:a_disagreement_analysis}.
% \begin{figure}[t!]
%     \centering
%     \includegraphics[width=\columnwidth]{Figures/ratio_of_culturespecific_posts.pdf}
%     \caption{Illustration of the two-step procedure of CREHate construction; 1) cultural post collection and 2) cross-cultural annotation. 
%     We show examples of how annotations on identical posts differ across countries.}
%     \label{fig:4_3_ratio_cp}
% \end{figure}
% \begin{figure}[t!]
%     \centering
%     \includegraphics[width=\columnwidth]{Figures/ratio_of_disagreement_cp.pdf}
%     \caption{Illustration of the two-step procedure of CREHate construction; 1) cultural post collection and 2) cross-cultural annotation. 
%     We show examples of how annotations on identical posts differ across countries.}
%     \label{fig:4_3_ratio_dis_cp}
% \end{figure}
% \begin{figure}[t!]
%     \centering
%     \includegraphics[width=\columnwidth]{Figures/ratio_of_disagreement_noncp.pdf}
%     \caption{Illustration of the two-step procedure of CREHate construction; 1) cultural post collection and 2) cross-cultural annotation. 
%     We show examples of how annotations on identical posts differ across countries.}
%     \label{fig:4_3_ratio_dis_noncp}
% \end{figure}

% \begin{figure}[t!]
% \subfloat[Culture-specific Posts]{
%   \includegraphics[clip,width=\columnwidth]{Figures/ratio_of_disagreement_cp.pdf}
% }

% \subfloat[Non-culture-specific Posts]{
%   \includegraphics[clip,width=\columnwidth]{Figures/ratio_of_disagreement_noncp.pdf}
% }

% \caption{main caption}
% \end{figure}y

\textit{Sarcasm} heightens challenges in intercultural agreement in hate speech annotation, as annotators' sensitivity to sarcasm may vary depending on the topic and the annotators' cultural backgrounds. 
Furthermore, sarcasm referring to a specific culture-specific context may be difficult for annotators from different backgrounds to accurately identify.

\textit{Personal bias} also plays a significant role in label disagreements, as they may arise when annotators hold differing opinions about specific topics, especially divisive issues.
For example, if the post is about divisive topics within the annotator's culture, their personal bias would have a larger impact on the annotation.

\textit{Swearing} is important in label disagreement since annotators' perceived offensiveness of a swear word can vary depending on their backgrounds.
Different cultures may have varying perceptions of swear words based on their usage and social context, resulting in label disagreements on the text containing them.

\textit{Not complete} indicates insufficient information for annotators to fully comprehend the post.
Annotators from diverse cultures may struggle to label posts involving cultural references or nuances from other cultures when crucial information is missing, requiring extra cultural background knowledge.

\paragraph{\textbf{Comparison between CC-SBIC and CP.}}
\label{sec:a_disagreement_analysis}
\begin{figure}[t!]
    \centering
    \includegraphics[width=\columnwidth]{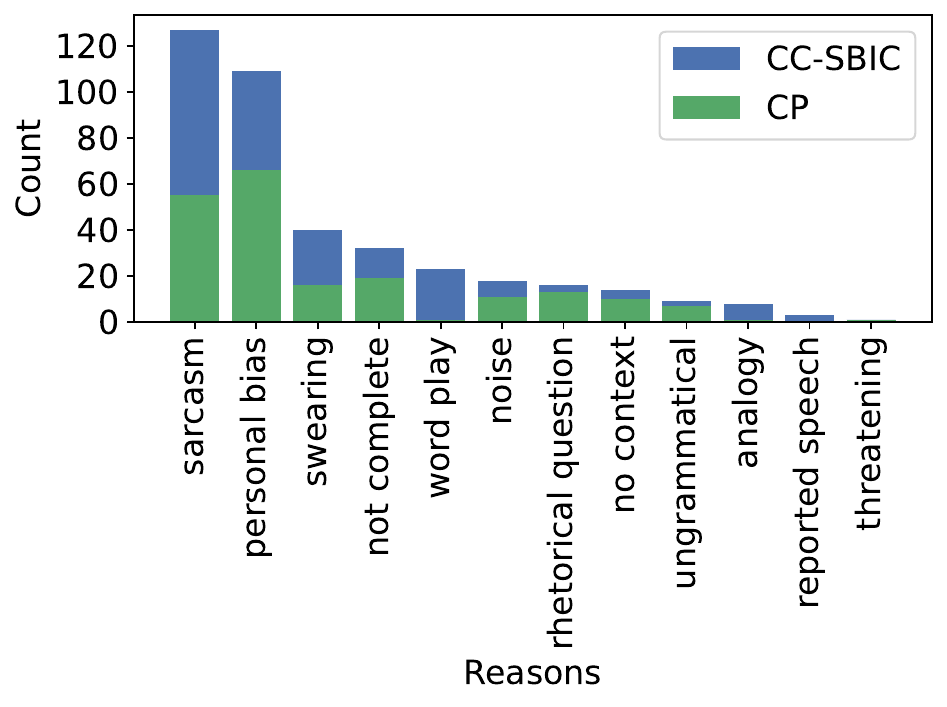}
    \caption{Disagreement reason count for CC-SBIC and CP posts.}
    \label{fig:disagreement_all}
\end{figure}
Figure \ref{fig:disagreement_all} shows the counts of all disagreement reason subtypes within CC-SBIC and CP posts. In both CC-SBIC and CP, \textit{sarcasm} and \textit{personal bias} are the two most significantly contributing reasons for label disagreements. However, there are some differences in the reasons for disagreement between the two dataset divisions. First of all, CP has more posts that the label disagreement is due to the \textit{personal bias} of annotators. This could be attributed to the comments on YouTube news videos included in CP, which primarily involve the authors' opinions on social issues handled within the videos. In addition, since CP posts contain more culturally intense topics within different countries in contrast to SBIC, they contain more \textit{not complete} posts, which require cultural knowledge for full comprehension. On the other hand, CC-SBIC has more posts containing \textit{word play} and \textit{swearing} compared to CP. One possible reason for this result is that people tend to be less constrained and write more freely on Twitter and hate sites, primary data collection sources not included in CP, compared to YouTube news comments.

\section{Experiments}
% Interpretation of hate speech varies depending on one's cultural background, highlighting the need to incorporate cultural diversity into the models used for hate speech classification.
This section evaluates the performance of current LLMs in hate speech classification on \crehate{}, with a specific focus on analyzing their performance with respect to country-specific annotations.

\noindent \textbf{Experimental Settings. }
We conduct zero-shot experiments using a multiple-choice question format.
We use five variations of the question prompt, each presented with two answer options: `Hate' and `Non-hate.'
We test both answer option orders for all prompts, resulting in 10 formats\thinspace\footnote{The prompts we use are listed in Appendix \ref{sec:app_prompts}.}.
All 1,580 instances of the CREHate corpus are used in the experiments, and all the prompts are generated by the authors.

The experiments are divided into two parts.
In \S\ref{sec:5_1_nopersona}, we ask the model to predict whether a given post is hate speech. 
In \S\ref{sec:5_2_wpersona}, we add a target country to the question prompt to examine the model's ability to make culture-specific predictions in hate speech detection.
For all results, we present the average values for all variations and perform a Wilcoxon rank-sum test to determine any statistically significant differences between each model's highest and lowest accuracy scores across different countries.

The models evaluated include GPT-4 (gpt-4-1106-preview), GPT-3.5 (gpt-3.5-turbo-1106), Orca 2 (7B)~\cite{orca2}, Flan-T5-XXL~\cite{flant5}, and OPT ~\cite{optiml}\thinspace\footnote{For GPT models, we use the OpenAI API and set the temperature as 0 to use greedy decoding (\url{https://platform.openai.com/docs/models}). For other models, we use the Huggingface Transformers~\cite{wolf-etal-2020-transformers} library (\url{https://github.com/huggingface/transformers}).}.

\subsection{Zero-shot Predictions and Country Labels}
\label{sec:5_1_nopersona}

% We assess whether hate speech classifications of LLMs exhibit any bias towards specific cultures, using prompts such as:
We obtain the predictions of LLMs using prompts such as:
\begin{mdframed}

Answer if this post is hate or not with a single alphabet letter among given answer choices a and b.%\\
% \\
% POST: \{POST\}\\
% a: \{Hate or Non-hate\}\\
% b: \{Non-hate or Hate\}\\
% answer:

\end{mdframed}
Table \ref{tab:5_llm} shows that the accuracy scores of all models, except for GPT models, are below 73\% when evaluated against all country labels on \crehate{}.
% All models other than GPT models demonstrate label similarities below 73\% against all country labels on CREHate as a whole. 
% Future work could involve leveraging CREHate to enhance the cultural awareness of these LLMs.
We also assess whether LLM-based hate speech predictions are biased towards particular cultures by comparing the accuracy of model predictions against the labels from each country.
Our analysis reveals that GPT models exhibit cultural bias, as their predictions display higher accuracy to the core Anglosphere cultures' labels.
While GPT-4 shows the highest overall accuracy across all country labels with an average value of 78.2\%, it also exhibits a significant performance gap with a maximum value of 6.79\%, most prominently between US labels (highest accuracy) and SG labels (lowest accuracy).
These findings suggest that high model accuracy does not necessarily equate to fairness, highlighting the need for more diverse training datasets and methods to mitigate cultural biases.

To determine if the IAA differences among countries (as shown in \S\ref{sec:4_2_label_agreement_country}) are the primary cause of varying accuracies, we examine the model accuracy on posts with unanimous annotator agreement within each country\thinspace\footnote{Please refer to Table \ref{tab:a_unanimous} for the results for all models in the Appendix.}.
Our analysis reveals that GPT-4 shows higher accuracy for US labels (95.25\%) and lower accuracy for SG labels (87.11\%), even on unanimously agreed posts. 
This suggests that the bias is inherent to the model's processing rather than a reflection of annotation quality.
% Given GPT-4's impressive performances across various tasks~\cite{openai2023gpt4}, it's crucial for researchers to consider its cultural biases. 

Furthermore, we observe that the overall accuracy for CP posts is lower than that of CC-SBIC across all countries.
Even for posts with unanimous annotator agreement within each country for the two dataset divisions, accuracies for CC-SBIC are higher than those on CP for most models.
This indicates difficulties in models classifying hate speech in CP posts explicitly sourced from countries other than the US.
% This can be due to several reasons, including the possibility of SBIC posts being part of the training data of the models as it was curated in 2020.
% \usepackage{booktabs}
\begin{table}[t!]
% \smCREHate
\resizebox{\columnwidth}{!}{
\begin{tabular}{@{}ll|ccccc@{}}
\toprule
\textbf{Model}           & \textbf{Data} & \textbf{GB}     & \textbf{US} & \textbf{AU} & \textbf{ZA} & \textbf{SG} \\ \midrule
\multirow{3}{*}{GPT-4}  & CREHate & 79.66           & \textbf{80.64*} & 78.02           & 78.03       & \underline{74.65}     \\
                        & \hspace{3mm}CC-SBIC & 80.74           & \textbf{82.13*} & 79.28           & 80.63       &\underline{75.34}     \\
                        & \hspace{3mm}CP      & 77.91           & \textbf{78.21*} & 75.96           & 73.79       &\underline{73.54}     \\ \midrule
\multirow{3}{*}{GPT-3.5} & CREHate       & \textbf{72.47*} & 70.62       & 72.39       &\underline{69.28} & 71.94       \\
                        & \hspace{3mm}CC-SBIC & 75.73           & 75.00           & \textbf{75.75*} &\underline{73.20} & 75.10           \\
                        & \hspace{3mm}CP      & \textbf{67.13*} & 63.47           & 66.90           &\underline{62.87} & 66.77           \\ \midrule
\multirow{3}{*}{Orca 2} & CREHate & \textbf{69.99}  & 69.09           & 69.80           & 68.80       &\underline{68.61}     \\
                        & \hspace{3mm}CC-SBIC & 72.19           & \textbf{72.58}  & 72.13           & 72.15       &\underline{70.87}     \\
                        & \hspace{3mm}CP      & \textbf{66.38}  & 63.38           & 65.98           &\underline{63.32} & 64.92           \\ \midrule
\multirow{3}{*}{Flan T5} & CREHate       & \textbf{68.58}  &\underline{67.49} & 68.28       & 68.35       & 68.15       \\
                        & \hspace{3mm}CC-SBIC & 72.49           & \textbf{72.86*} & 71.84           & 71.63       &\underline{70.35}     \\
                        & \hspace{3mm}CP      & 62.18           &\underline{58.72}     & 62.48           & 62.98       & \textbf{64.55*} \\ \midrule
\multirow{3}{*}{OPT}    & CREHate & 66.25           & \textbf{69.29}  & 64.68           & 66.94       &\underline{64.11}     \\
                        & \hspace{3mm}CC-SBIC & 65.22           & \textbf{68.75}  & 64.27           & 67.68       &\underline{63.06}     \\
                        & \hspace{3mm}CP      & 67.93           & \textbf{70.18*} &\underline{65.36}     & 65.72       & 65.83           \\ \bottomrule
\end{tabular}}
\caption{Accuracy of the models in terms of each country's labels in each dataset division. The highest score is highlighted in \textbf{bold}, while the lowest score is \underline{underlined}. The asterisk (*) means the two values differ significantly ($p < 0.05$).}
\label{tab:5_llm}
\end{table}

\subsection{Culture-Specific Hate Speech Classification}
\label{sec:5_2_wpersona}
In this section, we explore the culture-specific hate speech classification capability of GPT-4, which is the model that performed better overall in our previous experiment.
%as it show highest performance overall.
We aim to evaluate the model's ability to make culturally tailored predictions regarding hate speech classification using prompts that target specific countries:
\begin{mdframed}

Answer if this post is hate or not \textbf{in \{\textit{country}\}} with a single alphabet letter among given answer choices a and b.

\end{mdframed}

% Our analysis reveals that despite targeting specific countries in the prompts, GPT-4's performance does not significantly vary and still shows a preference for US and GB annotations (Table \ref{tab:5_llm_country}).
% Only 6.5-8\% of the label predictions differed compared to the predictions from the original prompt, not significantly affecting the accuracy scores for all country labels.
% This suggests that even a model known for its ability to follow instructions and achieve impressive results in other tasks cannot make culturally tailored hate speech detection by mentioning the target country.
Our analysis indicates that GPT-4's performance remains consistent, regardless of including specific country information in the prompts, as shown in Table \ref{tab:5_llm_country}. 
The inclusion of country names in the prompts led to only a marginal variation in predictions, with a 6.5-8\% difference from the predictions obtained using the original, non-country-specific prompts. 
This finding implies that providing country context alone does not significantly enhance GPT-4's ability to identify hate speech accurately across different cultural contexts. 
Consequently, this underscores a limitation in the model's capability to adapt its hate speech detection to specific cultural nuances merely through the explicit mention of a country in the prompt.
However, introducing cultural background information or other extra knowledge about the target country, or even using different prompts, may show different results.%., demonstrating that the model considers the cultural background and makes different predictions. 
We leave the exploration of prompt engineering that could enhance culture-specific hate speech detection in LLMs for future work.
% \usepackage{booktabs}
\begin{table}[t!]
\small
\centering

% \resizebox{\columnwidth}{!}{
\begin{tabular}{@{}cccccc@{}}
\toprule
\textbf{Prompt} & \textbf{GB}     & \textbf{US}     & \textbf{AU} & \textbf{ZA} & \textbf{SG} \\ \midrule
Original        & 79.66           & \textbf{80.64*} & 78.02       & 78.03       &\underline{74.65} \\ \midrule
+ in GB         & 79.66           & \textbf{80.28*} & 77.97       & 77.36       &\underline{73.52} \\
+ in US         & 79.27           & \textbf{80.26*} & 77.34       & 77.09       &\underline{73.32} \\
+ in AU         & \textbf{79.62*} & 79.59           & 77.95       & 77.40        &\underline{73.48} \\
+ in ZA         & 79.07           & \textbf{79.61*} & 77.38       & 77.44       &\underline{72.91} \\ 
+ in SG         & \textbf{79.70*} & 79.56           & 78.02       & 77.53       &\underline{73.27} \\ \bottomrule
\end{tabular}
% \begin{tabular}{@{}lcccccc@{}}
% \toprule
% \textbf{Model}           & \textbf{Prompt} & \textbf{AU} & \textbf{GB}    & \textbf{SG} & \textbf{US}    & \textbf{ZA} \\ \midrule
% \multirow{6}{*}{GPT-4}   & Original & 78.02       & 79.66          & \underline{74.65} & \textbf{80.64} & 78.03       \\
%  & + in AU & 77.95          & \textbf{79.62} & \underline{73.48} & 79.59          & 77.40       \\
%  & + in GB & 77.97          & 79.66          & \underline{73.52} & \textbf{80.28} & 77.36       \\
%  & + in SG & 78.02          & \textbf{79.70} & \underline{73.27} & 79.56          & 77.53       \\
%  & + in US & 77.34          & 79.27          & \underline{73.32} & \textbf{80.26} & 77.09       \\
%  & + in ZA & 77.38          & 79.07          & \underline{72.91} & \textbf{79.61} & 77.44       \\ \midrule
% \multirow{6}{*}{GPT-3.5} & Original        & 72.39       & \textbf{72.47} & 71.94       & 70.62          & \underline{69.28} \\
%  & + in AU & \textbf{72.23} & 71.76          & 71.03       & 69.53          & \underline{68.92} \\
%  & + in GB & \textbf{72.48} & 72.17          & 71.50       & 69.73          & \underline{69.10} \\
%  & + in SG & \textbf{72.42} & 72.36          & 71.49       & 70.05          & \underline{69.72} \\
%  & + in US & \textbf{73.49} & 73.22          & 72.23       & 71.54          & \underline{70.49} \\
%  & + in ZA & \textbf{72.42} & 72.20          & 71.42       & 70.37          & \underline{69.64} \\ \bottomrule
% \end{tabular}
% }
\caption{Accuracy of GPT-4 in terms of each country labels when asked to predict whether a given post is hateful within specific countries (e.g., ``\textit{Answer if this post is hate or not in Australia.}''). The highest score is highlighted in \textbf{bold}, while the lowest score is \underline{underlined}. The asterisk (*) means the two values differ significantly ($p < 0.05$).}
\label{tab:5_llm_country}
\end{table}

% \section{Culturally-adapted Models}
% \input{Content/5_Culturally_Adapted_Models}

\section{Conclusion}

In this paper, we analyze how cultural differences across English-speaking countries affect hate speech annotations. 
To this end, we develop \crehate{}, a cross-cultural English hate speech dataset comprising 1,580 posts from five English-speaking countries---AU, GB, SG, US, and ZA.
Our work shows that there are notable variations in hate speech interpretations between these countries through various statistical methods. 
The overall agreement on hate speech identification across all countries is only 56.2\%, with an average pairwise country disagreement of 21.2\%. 
Qualitative analysis suggests these differences stem from varied understandings of sarcasm and annotators' biases on divisive topics. 
We also discover that GPT models display higher accuracies with labels from Anglosphere cultures and fail to make culturally tailored predictions when the target country is given.

This research establishes a foundational framework for continuously evaluating and adapting hate speech models and datasets. 
We suggest expanding \crehate{} to include more countries and posts to create a comprehensive tool for assessing cultural biases in model predictions and enhancing culturally tailored hate speech detection. 
We urge collaborative efforts in constructing datasets with broad cultural references and contextual nuances.
Annotators with relevant cultural knowledge should be employed to construct a more representative cultural dataset. 
Such a comprehensive approach is crucial for developing more effective, culturally sensitive hate speech classifiers and promoting safer and more inclusive online communication.

Current hate speech detection tools often struggle with cultural biases, particularly skewing toward US or Western perspectives. This results in inadequate representation and understanding of sociocultural contexts from other English-speaking countries. For example, a phrase considered derogatory in one culture might be benign in another, leading to false positives or negatives in detection. Interpretations of posts can vary significantly between cultures, influenced by factors like local idioms, societal norms, and historical contexts. Therefore, a more nuanced and inclusive approach to dataset creation and algorithm development is needed to ensure broader cultural representation and sensitivity.

\section*{Limitations}
CREHate consists of 1,580 posts, making it relatively small compared to other existing English hate speech datasets.
Moreover, the collection of culture-specific posts was limited to Reddit and YouTube based on fixed hate-related keywords, which may introduce bias into the collected posts.
Also, employing a single crowdsourcing platform for collecting each country's annotation may lead to annotator bias, as different platforms possess varying user demographics.
To enhance the representativeness and generalizability of our findings, we anticipate future efforts to expand our dataset by using diverse platforms and post collection methods.

Considering that many countries are multicultural, it is also essential to examine within-country annotation differences.
For instance, Singapore has a diverse population, including Chinese, Malaysians, and Indians.
Exploring hate speech annotation differences across different ethnicities within a country presents another avenue for investigation. Moreover, although we recruit annotators from countries where English is one of their official language(s), this may not be enough to cover all English-speaking cultures.
Further study is needed to include English as a Foreign Language (EFL) learners in cross-cultural hate speech detection.
Moreover, the same approach could be extended to languages other than English (e.g., Spanish) spoken in various countries.

There are other subjective tasks that are affected by cultural context, such as common sense reasoning.
Future research could extend the scope of our study to other tasks by constructing datasets tailored towards specific cultures, both within and across countries with diverse languages.

\section*{Ethics Statement}
% This research was conducted with full approval from the Institutional Review Board (IRB).
This research project was performed under approval from KAIST IRB (KH2023-068).
The instructions that were given to the annotators, including the disclaimer, can be seen in Figure \ref{fig:ann_disclaimer} in the Appendix.
We made sure to inform the annotators from the crowdsourcing platforms that the contents they encounter during the annotation task may potentially be offensive or distressing.
We also provided access to online therapy platforms and encouraged the annotators to seek help in case they experience any strong negative reactions or mental distress.

We conducted our crowd worker recruitment without any discrimination based on age, ethnicity, disability, or gender. 
Our workers are compensated at a rate higher than Prolific's ethical standards.
Our payment principles are based on the ethical standards of Prolific, and we ensure that our workers are compensated at a rate higher than the minimum wage of £9.00 per hour. 
It is worth noting that this amount exceeds the federal minimum wage in the United States and Singapore, where the annotation process was held on other crowdsourcing platforms.

We are aware of the potential risk involved in releasing a dataset containing hate speech or offensive language.
We will explicitly state the terms of usage, emphasizing our unequivocal disapproval of any form of malicious exploitation.
We urge researchers and practitioners to harness this dataset only for constructive purposes. 
We expect our dataset to contribute to developing more equitable and culturally sensitive automated content moderation systems.
We emphasize our unequivocal disapproval of any form of malicious exploitation of our dataset, including any misuse of our dataset for generating hateful language.
We demand that researchers and practitioners use this dataset solely for constructive purposes.

We used AI assistants --- ChatGPT\thinspace\footnote{\url{https://chat.openai.com}}, Google Translate\thinspace\footnote{\url{https://translate.google.com}}, and Grammarly\thinspace\footnote{\url{https://app.grammarly.com}} --- to assist with editing and translating sentences in our paper writing.
The \crehate{} dataset is licensed under CC BY-SA 4.0.

\section*{Acknowledgement}
This project was funded by the KAIST-NAVER hypercreative AI center. Alice Oh is funded by Institute of Information communications Technology Planning Evaluation (IITP) grant funded by the Korea government(MSIT) (No. 2022-000184, Development and Study of AI Technologies to Inexpensively Conform to Evolving Policy on Ethics).
Jose Camacho-Collados is supported by a UKRI Future Leaders Fellowship.

% % Entries for the entire Anthology, followed by custom entries
\bibliography{anthology,custom}
\section*{Appendix}
\appendix

\section{Dataset Construction Details}
\subsection{Post Collection}
\label{sec:A_1_post_collection}
\subsubsection{SBIC}
\label{sec:A_1_1_sbic}
\begin{table}[ht!]
\centering
\small
% \resizebox{\columnwidth}{!}{
\begin{tabular}{@{}c|c|c@{}}
\toprule
\textbf{Category} & \multicolumn{1}{l|}{\textbf{\begin{tabular}[c]{@{}c@{}}Original\\ Test Set (\%)\end{tabular}}} & \textbf{\begin{tabular}[c]{@{}c@{}}Sampled\\ Dataset (\%)\end{tabular}}  \\ \midrule
Race/Ethnicity    & 819 (17.5)   & 150 (15.3)                                                               \\
Gender/Sexuality  & 503 (10.7)   & 150 (15.3)                                                              \\
Religion/Culture  & 495 (10.6)   & 150 (15.3)                                                              \\
Victims           & 215 (4.6)   & 150 (15.3)                                                                \\
Disability        & 112 (2.4)   & 112 (11.4)                                                               \\
Social/Political  & 104 (2.2)   & 104 (10.6)                                                              \\
Body/Age          & 58 (1.2)  & 58 (5.9)                                                             \\ \midrule
Non-hate          & 2765 (58.9)  & 327 (33.4)                 \\ \midrule
\textbf{Total}    & 4691         & 980                                                               \\ \bottomrule
\end{tabular}
% }
\caption{Category distribution within the original and the sampled SBIC test set. CC-SBIC posts are comprised of randomly sampled 980 posts from the original SBIC test set, maintaining balance among target group categories. Multi-labeled group categories are split into multiple individual categories when counting.}
\label{tab:a_1_sbic_data_distribution}
\end{table}
Posts in SBIC originate from subReddits, microaggressions corpus~\cite{breitfeller-etal-2019-finding}, Twitter~\cite{founta2018large,Davidson_Warmsley_Macy_Weber_2017,waseem-hovy-2016-hateful-tacl}, and hate sites (Gab\thinspace\footnote{\url{https://files.pushshift.io/gab/GABPOSTS_CORPUS.xz}} and Stormfront~\cite{de-gibert-etal-2018-hate}).
The dataset contains offensive posts targeted towards diverse demographic group categories, including \textit{race/ethnicity}, \textit{gender/sexuality}, \textit{religion/culture}, \textit{victims}, \textit{disability}, \textit{social/political}, and \textit{body/age}.
We maintain a 2:1 ratio between hateful and non-hateful posts in our sampled SBIC data to prioritize our analysis on hate speech rather than non-hate speech.
The sampled SBIC data statistics are shown in Table \ref{tab:a_1_sbic_data_distribution}.

\subsubsection{Cultural Posts}
\label{sec:A_1_1_cultural_posts}
\begin{table}[t!]
\centering
% \small
\resizebox{\columnwidth}{!}{
\begin{tabular}{@{}c|c|c@{}}
\toprule
\textbf{ Source} & \multicolumn{1}{c|}{\textbf{Reddit}}                                                                                        & \multicolumn{1}{c}{\textbf{YouTube}}                              \\ \midrule
{ AU}               & \begin{tabular}[c]{@{}c@{}}r/australia, r/Australian, r/melbourne,\\ r/sydney, r/perth, r/brisbane, r/Adelaide\end{tabular} & Sky News Australia                                                \\ \midrule
{ GB}               & \begin{tabular}[c]{@{}c@{}}r/unitedkingdom, r/CasualUK, r/england, \\ r/Scotland, r/Wales, r/northernireland\end{tabular}   & \begin{tabular}[c]{@{}l@{}}SkyNews, GBNews\end{tabular}        \\ \midrule
{ SG}               & \begin{tabular}[c]{@{}c@{}}r/singapore, r/SingaporeRaw, \\ r/singaporehappenings, r/singapuraa\end{tabular}                 & \begin{tabular}[c]{@{}l@{}}CNA, The Straits Times\end{tabular} \\ \midrule
{ ZA}               & \begin{tabular}[c]{@{}c@{}}r/southafrica, r/RSA, r/capetown, \\ r/johannesburg, r/Durban, r/Pretoria\end{tabular}           & \begin{tabular}[c]{@{}l@{}}SABC News, eNCA\end{tabular}        \\ \bottomrule
\end{tabular}}
\caption{Data sources for each country. We crawled comments from country-specific subreddits and news platforms' YouTube channels.}
\label{tab:3_1_2_post_source}
\end{table}
\begin{table}[t!]
\centering
\resizebox{\columnwidth}{!}{
\begin{tabular}{@{}l|ccccc@{}}
\toprule
\multicolumn{1}{c|}{}             & \textbf{AU} & \textbf{GB} & \textbf{US} & \textbf{SG} & \textbf{ZA} \\ \midrule
\textbf{No. of Annotators}        & 216         & 405         & 166          & 103         & 173         \\ \midrule
\textbf{Gender} (\%)                  &             &             &             &             &             \\
\hspace{3mm}male                  & 51.39      & 45.68      & 53.61      & 54.46      & 50.29      \\
\hspace{3mm}female                & 46.30      & 52.35      & 46.39      & 44.55      & 48.55      \\
\hspace{3mm}non-binary            & 2.31       & 1.98       & -           & 0.99       & 1.16       \\ \midrule
\textbf{Race} (\%)                &             &             &             &             &             \\
\hspace{3mm}Asian                 & 23.61      & 4.20       & 4.22       & 100.00     & 4.05       \\
\hspace{3mm}Black                 & 0.46       & 2.72       & 6.63       & -           & 77.46      \\
\hspace{3mm}Hispanic              & -           & 0.25       & 0.60       & -           & -           \\
\hspace{3mm}Middle Eastern        & 1.85       & 0.25       & 0.60       & -           & 0.58       \\
\hspace{3mm}White                 & 67.59      & 89.14      & 86.75      & -           & 11.56      \\
\hspace{3mm}Other                 & 6.49       & 3.44       & 1.20       & -       & 6.35       \\ \midrule
\textbf{Level of Education} (\%)          &             &             &             &             &             \\
\hspace{3mm}Below High School & 1.39       & 0.74       & -           & -           & -           \\
\hspace{3mm}High School           & 11.11      & 14.07      & 16.87      & 15.84      & 16.76      \\
\hspace{3mm}College               & 20.83      & 23.70      & 36.14      & 15.84      & 28.90      \\
\hspace{3mm}Bachelor              & 46.30      & 43.95      & 40.96      & 62.38      & 48.55      \\
\hspace{3mm}Master’s Degree  & 17.59      & 15.80      & 4.82       & 5.94       & 5.78       \\
\hspace{3mm}Doctorate       & 2.78       & 1.73       & 1.20       & -           & -           \\ \midrule
\textbf{Age} (\%)                      &             &             &             &             &             \\
\hspace{3mm}18-19                 & 2.31       & 1.73       & -           & 1.98       & 0.58       \\
\hspace{3mm}20-29                 & 52.31      & 27.90      & 3.01       & 60.40      & 73.41      \\
\hspace{3mm}30-39                 & 22.22      & 27.90      & 41.57      & 27.72      & 18.50      \\
\hspace{3mm}40-49                 & 15.28      & 21.73      & 25.90      & 2.97       & 2.89       \\
\hspace{3mm}50-59                 & 4.63       & 13.58      & 18.07      & 1.98       & 2.89       \\
\hspace{3mm}60-69                 & 2.31       & 5.43       & 9.04       & 4.95       & 1.73       \\
\hspace{3mm}70-79                 & 0.46       & 1.73       & 2.41       & -           & -           \\
\hspace{3mm}80-89                 & 0.46       & -           & -           & -           & -           \\ \midrule

\textbf{Political Orientation} (\%)    &             &             &             &             &             \\
\hspace{3mm}Liberal/Progressive   & 42.59      & 29.88      & 39.76      & 15.84      & 21.97      \\
\hspace{3mm}Moderate Liberal      & 27.78      & 29.38      & 22.89      & 19.80      & 19.08      \\
\hspace{3mm}Independent           & 18.52      & 17.53      & 11.45      & 37.62      & 35.26      \\
\hspace{3mm}Moderate Conservative & 6.02       & 14.07      & 16.27      & 14.85      & 14.45      \\
\hspace{3mm}Conservative          & 3.70       & 5.68       & 9.04       & 9.90       & 9.25       \\
\hspace{3mm}Other                 & 1.39       & 3.46       & 0.60       & 1.99       & -       \\ \midrule
\textbf{Religion} (\%)                 &             &             &             &             &             \\
\hspace{3mm}None                  & 64.81      & 62.47      & 50.60      & 38.61      & 16.19      \\
\hspace{3mm}Christian             & 20.83      & 28.89      & 37.95      & 26.73      & 75.72      \\
\hspace{3mm}Buddhism              & 2.78       & 0.74       & -           & 24.75      & -           \\
\hspace{3mm}Islam                 & 0.93       & 3.21       & 0.60       & 4.95       & 3.47       \\
\hspace{3mm}Judaism               & -           & 0.49       & 1.81       & 1.98           & -           \\
\hspace{3mm}Hinduism              & 0.46       & 0.25       & -           & -           & -           \\
\hspace{3mm}Irreligion            & 5.56       & 1.23       & 3.61       & -           & 0.58       \\
\hspace{3mm}Other                 & 4.63       & 2.72       & 5.42       & 2.98       & 4.04       \\ \bottomrule

\end{tabular}
}
\caption{Annotator demographic statistics from each country. }
\label{tab:a_annotator_demographic}
\end{table}
Specific subReddits and news sites used for post crawling are shown in Table \ref{tab:3_1_2_post_source}.
There is only one news site for Australia, as no other YouTube channels of news sites provided by the workers allow comments.
On Reddit, we extract all comments on the posts that include the target group names or the keywords provided by workers. 
On YouTube, we search using the query, `<media name> + <target group name>', to locate comments related to the target groups (e.g., `BBC news pakistani’).
We only include comments and posts written in 2020 or later for an up-to-date dataset.

After crawling cultural posts from the four countries, we go through a pre-annotation stage in order to balance hate and non-hate speech in our dataset to the extent possible.
The process begins by randomly selecting 300 comments from each country, balancing those from Reddit and YouTube.
We then obtain two annotations per comment from the source country of the comments.
Subsequently, we curate a collection of 150 comments by selecting 50 from each of the three hate annotation counts, ranging from 0 to 2.
With this procedure, we get 600 cultural posts from four countries.

\subsubsection{Post-processing of Posts}
SBIC posts and crawled Reddit and YouTube comments contained usernames and URLs that were not masked. 
To anonymize all posts, we mask the usernames as \verb|@USER|, and URLs as \verb|URL|.

\subsubsection{Terms of Use}
Our research is performed in the public interest under GDPR, as we meet the substantial public interest conditions as academic research.
The SBIC dataset is licensed under CC BY 4.0. 
We use Reddit's official data API, following the terms of use mentioned in `Data API Terms'\thinspace\footnote{\url{https://www.redditinc.com/policies/data-api-terms}}.
We use YouTube API from Google for Developers site, following the terms of use mentioned in Complying with YouTube's Developer Policies page\thinspace\footnote{\url{https://developers.google.com/youtube/terms/developer-policies-guide?hl=en}}.
The \crehate{} dataset is licensed under CC BY-SA 4.0.

\subsection{Annotator Demographics}
% \paragraph{}
Table \ref{tab:a_annotator_demographic} shows the total number of annotators and the proportion of all demographic groups among annotators from each country. 
The first three demographic categories---gender, ethnicity, and level of education---are shown to be factors that significantly affect hateful post annotations. 

\subsection{Annotation Process}
\label{sec:quality_control}
\begin{figure*}[t!]
    \centering
    \includegraphics[width=0.80\textwidth]{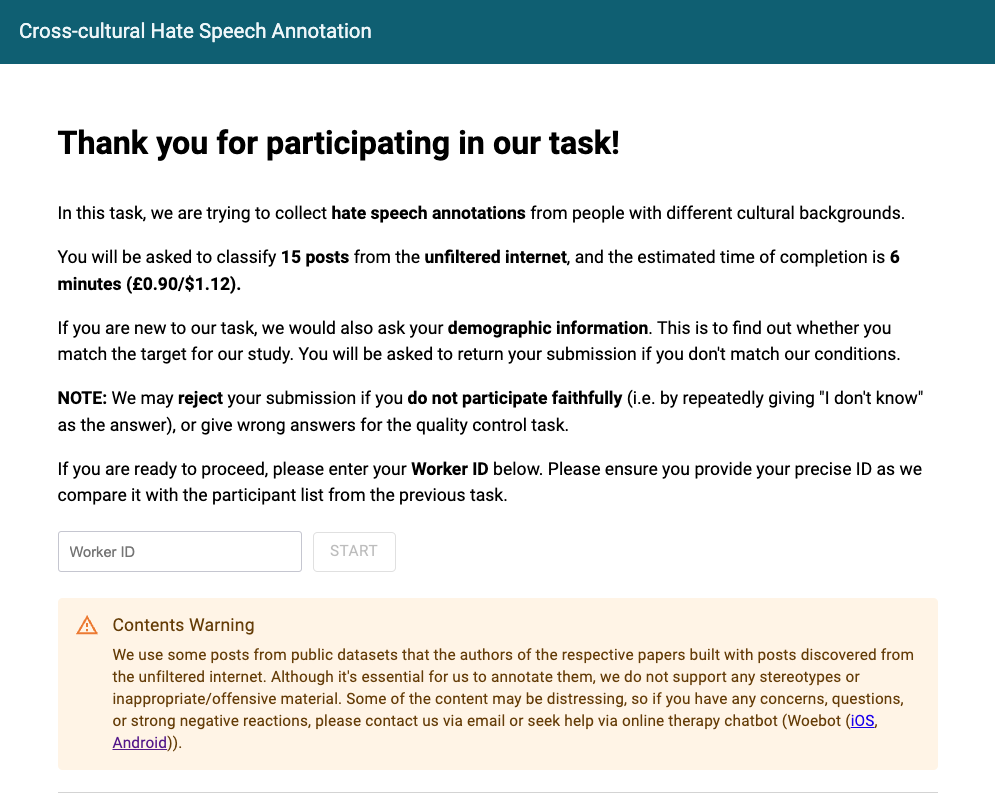}
    \caption{Disclaimer and instruction shown to the annotators.}
    \label{fig:ann_disclaimer}
\end{figure*}
\begin{figure*}[t!]
    \centering
    \includegraphics[width=0.85\textwidth]{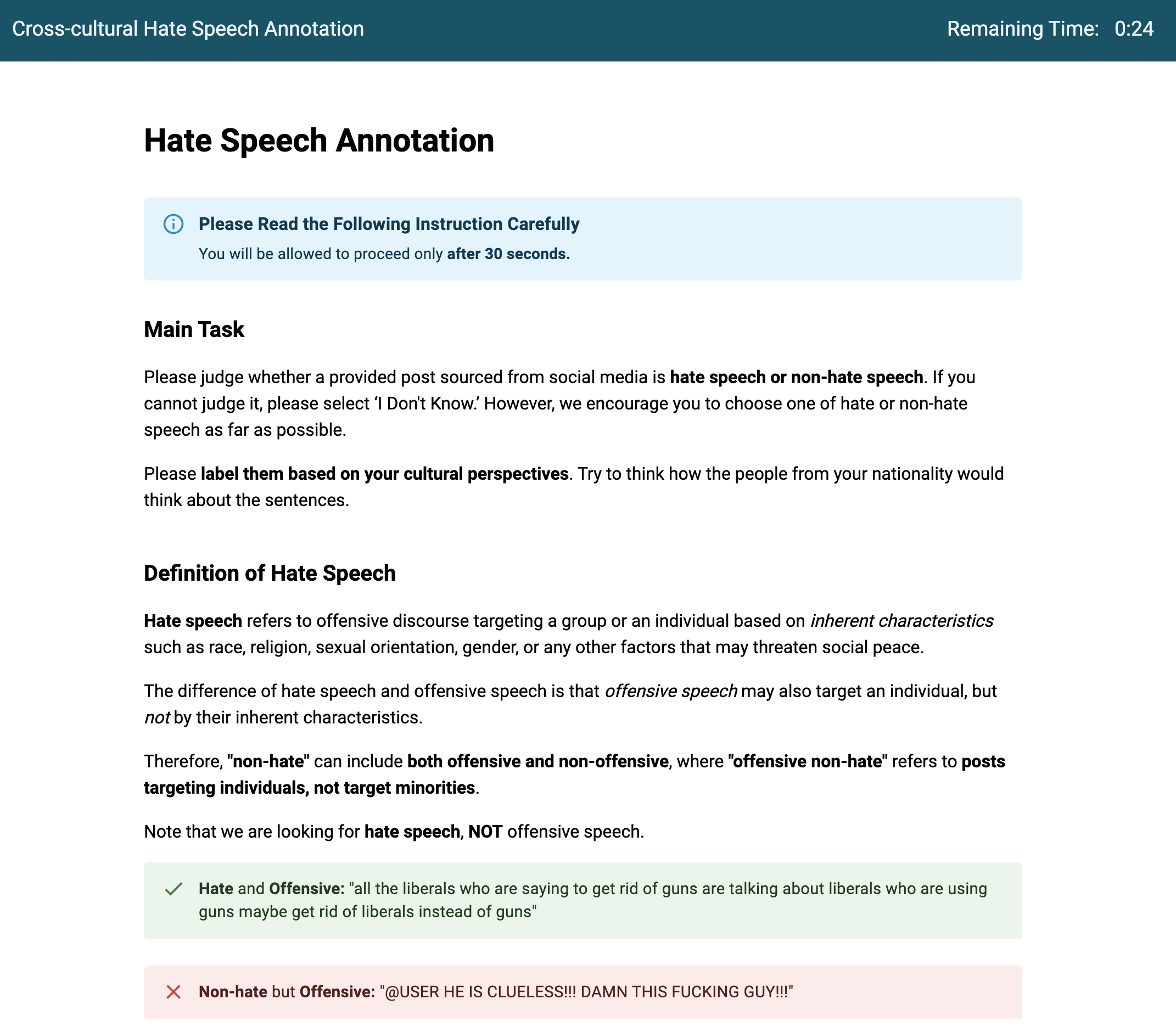}
    \caption{Guideline page of the hate speech annotation task shown to the annotators.}
    \label{fig:ann_guideline}
\end{figure*}
\begin{figure*}[t!]
    \centering
    \includegraphics[width=0.65\textwidth]{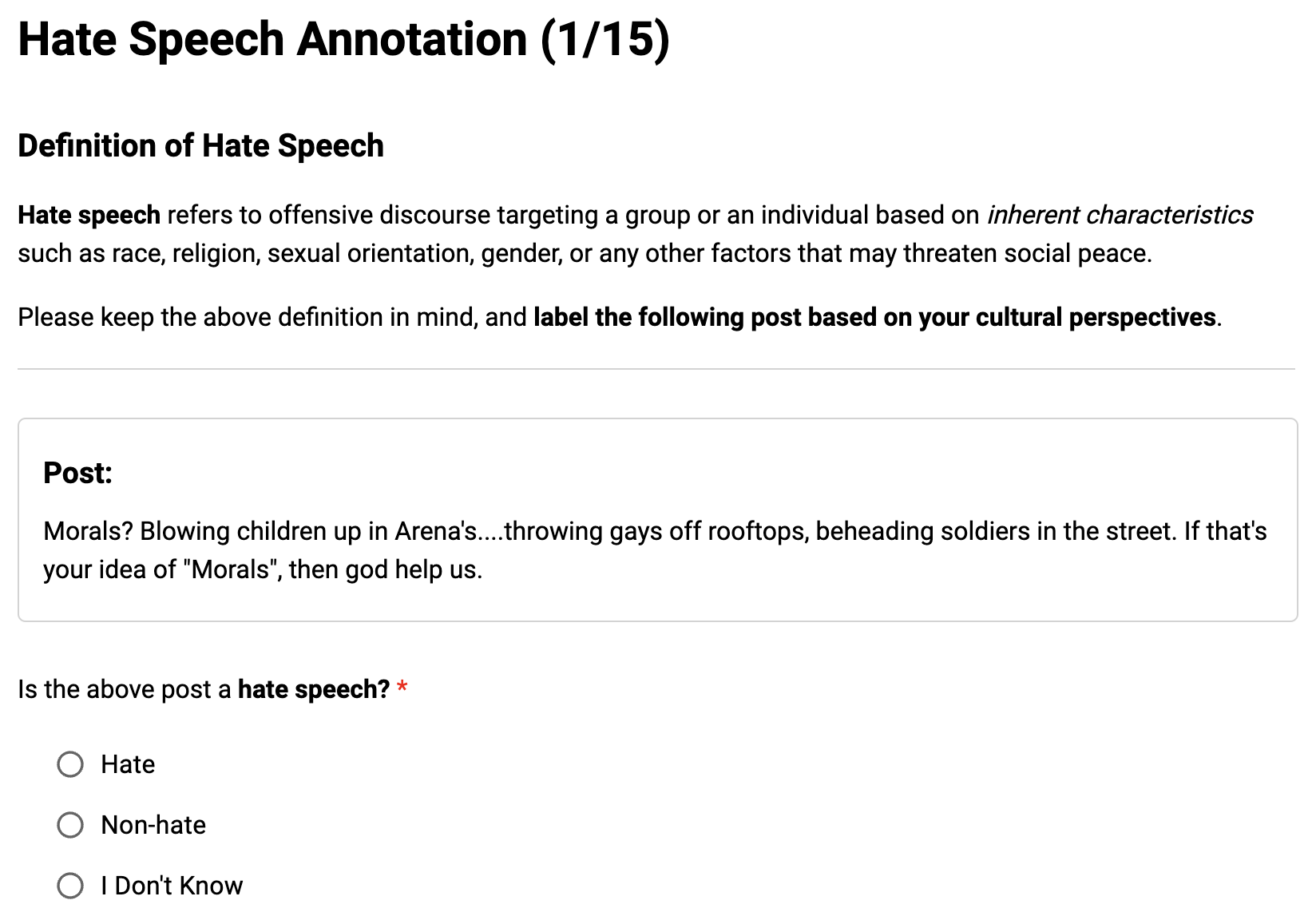}
    \caption{Hate speech annotation page shown to the annotators.}
    \label{fig:ann_page}
\end{figure*}

% \subsection{Annotation Quality Control}
Disclaimer and instruction are first shown to the annotators, as shown in Figure \ref{fig:ann_disclaimer}.
Each annotator is then asked to answer a demographic survey. If the annotator matches our target group mentioned in Section \S\ref{sec:3_1_2_Collecting_Cultural_Samples}, annotators proceed to the guideline page shown in Figure \ref{fig:ann_guideline}. After reading the guideline for a minimum of 30 seconds, annotators are asked to annotate 15 posts (Figure \ref{fig:ann_page}) that are randomly assigned among the remaining ones. 

We include two explicit and two implicit attention check questions among the annotation questions to ensure the dataset's quality. 
% The annotators are informed that they will be annotating a total of 15 samples, which include the implicit attention check questions.
The implicit attention check questions are selected from the samples on which all annotators from all countries agree in previously completed annotations. 
For the first round of the actual survey, we choose samples with total agreement from the pilot study. 
As the study progresses, we update them with the new samples the annotators agreed on. 
The two explicit attention checks instruct the annotators to choose a specific label. 
Only annotations from annotators that pass all attention checks are included in the dataset. 
To avoid a single annotator significantly affecting the annotation, each annotator can only contribute to a maximum of 5\% of the total annotation.

\section{Analysis on \textit{I Don't Know} Labels}
\label{sec:A_2_idk_analysis}
\begin{table}[t]
\centering
\small
\resizebox{\columnwidth}{!}{
% \begin{tabular}{@{}l|ccccc@{}}
% \toprule
% \multicolumn{1}{c|}{} & \textbf{AU} & \textbf{GB} & \textbf{SG} & \textbf{US} & \textbf{ZA} \\ \midrule
% \textbf{CREHate}      & 0.4373      & 0.4810      & 0.4310      & 0.1854      & 0.4019      \\ \midrule
% \hspace{3mm}\textbf{SBIC}         & 0.3388      & 0.3847      & 0.4918      & 0.1867      & 0.3694      \\
% \hspace{3mm}\textbf{CP}           & 0.5983      & 0.6383      & 0.3317      & 0.1833      & 0.4550      \\ \midrule
% \hspace{3mm}\textbf{CP$_{AU}$}       & 0.3133      & 0.5000      & 0.3467      & 0.1733      & 0.5667      \\
% \hspace{3mm}\textbf{CP$_{GB}$}       & 0.3933      & 0.3133      & 0.2267      & 0.0533      & 0.4400      \\
% \hspace{3mm}\textbf{CP$_{ZA}$}       & 1.0667      & 0.9267      & 0.5400      & 0.2133      & 0.2933      \\
% \hspace{3mm}\textbf{CP$_{SG}$}       & 0.6200      & 0.8133      & 0.2133      & 0.2933      & 0.5200      \\ \bottomrule
% \end{tabular}
\begin{tabular}{@{}l|ccccc@{}}
\toprule
\multicolumn{1}{c|}{} & \textbf{AU} & \textbf{GB} & \textbf{SG} & \textbf{US} & \textbf{ZA}\\ \midrule
\textbf{CREHate} & 0.0630 & 0.0678 & 0.0628 & 0.0273 & 0.0582       \\ \midrule
\hspace{3mm}\textbf{CC-SBIC} & 0.0504 & 0.0552 & 0.0712 & 0.0281 & 0.0532 \\
\hspace{3mm}\textbf{CP}      & 0.0835 & 0.0885 & 0.0491 & 0.0260 & 0.0663      \\ \midrule
\hspace{3mm}\textbf{CP$_{AU}$}  & 0.0482 & 0.0749 & 0.0502 & 0.0265 & 0.0838      \\
\hspace{3mm}\textbf{CP$_{GB}$}  & 0.0578 & 0.0498 & 0.0362 & 0.0086 & 0.0675      \\
\hspace{3mm}\textbf{CP$_{ZA}$}  & 0.1397 & 0.1252 & 0.0762 & 0.0337 & 0.0425      \\
\hspace{3mm}\textbf{CP$_{SG}$}  & 0.0883 & 0.1042 & 0.0338 & 0.0355 & 0.0713     \\ \bottomrule
\end{tabular}
}
\caption{The average ratio of \textit{I don't know} labels per post within each dataset division. 
}
\label{tab:A_avg_idk_cnt}
\end{table}

Table \ref{tab:A_avg_idk_cnt} reveals that the annotations in \crehate{} dataset contain only a few \textit{I don't know} labels. 
Across all countries, the ratio of \textit{I don't know} labels per post is only around 5\% within \crehate{}. 
Notably, annotations from the US exhibit a lower average \textit{I don't know} count compared to other countries.
% which may be attributed to the characteristics of the annotators. 
We collect annotations from the US using the Amazon MTurk, limiting participation to Masters. 
As highly experienced annotators, Masters may have refrained from selecting \textit{I don't know} labels. 
Additionally, for CP posts, there is a moderate tendency among annotators to select fewer \textit{I don't know} labels within posts originating from their own country.

% \label{sec:A_2_idk_analysis}
% \begin{figure}[t!]
%     \centering
%     \includegraphics[width=\columnwidth]{Figures/post_w_idk_ratio.pdf}
%     \caption{Ratio of posts with at least one \textit{I don't know} label by the ratio of \textit{hate} labels per post for each of the countries. }
%     \label{fig:post_w_idk_ratio}
% \end{figure}

We also analyze the correlation between the existence of \textit{I don't know} label and the ratio of \textit{hate} labels within posts.
Posts with disagreement among annotators from the same country, those with hate label ratios ranging from 0.2 to 0.8, tend to have a higher percentage of posts containing \textit{I don't know} labels.
% By calculating the ratio of posts with at least one \textit{I don't know} label among posts showing different ratios of \textit{hate} labels, we could see that posts that were more subjective, that is, posts that had some disagreement within the annotators from the same country, with hate ratios ranging from 0.2 to 0.8, tended to have higher ratio of posts containing \textit{I don't know} labels than those that had total agreement among the 5 annotators.
On the other hand, strongly hateful posts, where all annotators agreed that the post is hateful, tend to have fewer \textit{I don't know} labels, even compared to posts with annotators' unanimous agreement on annotating them as non-hate.
% As shown in Table \ref{tab:4_jp_f1}, the average Hate-F1 scores between countries were lower than average Non-hate-F1 scores.
This suggests that people tend to be more confident in labeling posts as hate, while they feel less confident about non-hateful posts.
% we may say that based on the fact that people were comparatively more certain in labeling posts as hate, the tendencies of people from different cultural backgrounds in labeling posts as hate differs more than non-hateful posts, which they feel less certain.

\section{Analysis on Pairwise Country Labels}
\label{sec:c_pairwise_label_analysis}

\begin{table}[t!]
\centering
\small
% \resizebox{\columnwidth}{!}{
\begin{tabular}{cc}
\toprule
\textbf{\begin{tabular}[c]{@{}c@{}}Country\\ Pairs\end{tabular}} & \textbf{\begin{tabular}[c]{@{}c@{}}Cultural\\ Distance\\ Index\end{tabular}} \\ \midrule
AU-SG                                                            & 3.842                                                                        \\
SG-US                                                            & 3.653                                                                        \\
GB-SG                                                            & 3.484                                                                        \\
SG-ZA                                                            & 2.178                                                                        \\
GB-ZA                                                            & 0.458                                                                        \\
GB-US                                                            & 0.446                                                                        \\
ZA-US                                                            & 0.344                                                                        \\
AU-ZA                                                            & 0.344                                                                        \\
AU-GB                                                            & 0.144                                                                        \\
AU-US                                                            & 0.015                                                                        \\ \bottomrule
\end{tabular}

% }
\caption{Cultural distance index values between country pairs~\cite{kogut1988effect,hofstede1984culture}.}
\label{tab:c_cultural_distance_index}
\end{table}
Table \ref{tab:c_cultural_distance_index} shows the cultural distance index values between all country pairs. 
Cultural distance index values tend to be higher in country pairs with Singapore, whereas those between core Anglosphere countries tend to be lower.
% The results from the McNemar Test~\cite{McNemar1947} further reveal more pronounced differences within CP than in CC-SBIC posts. 
% Specifically, CP$_{AU}$, CP$_{GB}$, and CP$_{ZA}$ exhibit significant differences in eight or more country pairs out of ten.
% CP$_{SG}$ shows significant differences in six out of ten country pairs. 
% However, in CC-SBIC posts, significant differences are observed in only two pairs: US and ZA, and AU and SG.
% These outcomes demonstrate significant variations in annotations among individuals from diverse countries, particularly within culturally specific posts.

\section{Disagreement Reason Taxonomy}
\label{sec:A_D_disagreement_reason}

As mentioned in Section \S\ref{sec:4_3_disagreement_analysis}, we leverage the taxonomy of annotation disagreement in subjective tasks from \citet{sandri-etal-2023}. The categories and subtypes of the taxonomy are shown in Table~\ref{tab:4_3_disagreement_types}.

\subsection{Category Definitions}
The definitions (taken from \citet{sandri-etal-2023}) and examples of each category are as follows.
\subsubsection{Sloppy Annotation}
\paragraph{Noise} corresponds to posts that are clearly labeled incorrectly, such as by annotating the following as hate speech: \textit{Blue Bell ice cream has one showing five kids one black playing in the fields and barn.}

\subsubsection{Ambiguity}
\paragraph{Analogy} encompasses comparison mechanisms like simile and metaphor. An analogy can cause ambiguity, especially for posts from different countries, as certain comparisons may only be used and understood in specific cultural groups. (e.g., \textit{Black people are like mitochondria They're the powerhouse of the cell}).
    
\paragraph{False Assertion} refers to instances where users convey opinions opposite to their actual beliefs or express falsehoods and exaggerations in relation to the context. (e.g., \textit{Another attempt backfired on them, George Floyd cured Covid-19 and opened up the economy!}).

\paragraph{Rhetorical Question} includes posing a question not with the intention of receiving an answer but rather to rhetorically highlight a concept (e.g., \textit{I read recently in cold black and white print that there was around 10,000 nigerians in Ireland ... Now you say close to 300,000 ????!}).
    
\paragraph{Sarcasm} is defined by the use of words to convey meanings opposite to their literal definitions, primarily employed for the purpose of mocking a particular topic or an individual (e.g., \textit{I joined a suicide cult today It’s known as the American Public School System}).

\paragraph{Word Play} is a figure of speech that utilizes literary elements to modify words, serving as a demonstration of someone's wit and cleverness. (e.g., \textit{I hate jokes about 9/11 victims they always fall flat}).

\paragraph{Reported Speech} is defined as the presentation of text that claims to be from a previous occasion and may have originated from another author. (e.g., \textit{Now on Britains pressing issues: can occasion cards be too offensive? Find out after we bubble wrap this child.}).

\subsubsection{Missing Information}

\paragraph{Ungrammatical} posts lead to disagreement because of noticeable language use, such as non-standard varieties, slang, code-switching, or simple typing errors. This includes posts containing the colloquial forms of English used in different countries, such as Singaporean English or South African English. (e.g., \textit{@USER chibai you eat babi at hawker centre, they got put headcover anot}).

\paragraph{No Context} encompasses a linguistic phenomena that require context for unequivocal interpretation, such as anaphora and deixis. Posts including URLs also fall into this category. (e.g., \textit{What's she got that stick for if she can read [that book]?}).

\paragraph{Not Complete} includes all instances that fail to convey complete information and do not fit into any other subcategory. This subcategory also includes posts that require additional cultural common-sense knowledge to understand. (e.g., \textit{You think the ANC would have sympathy with not wanting to be caledon a terrorist organisation}).

\subsubsection{Subjectivity}

\paragraph{Personal Bias} includes post that belong to divisive topics like politics, COVID-19, social movements, etc. Although it is challenging to precisely identify disagreements stemming from personal bias when annotators are unknown, posts within these contentious topics are likely to be inferred as containing personal bias due to the heightened probability. (e.g., \textit{You’re Irish and you support your pals the Provisional IRA murderers of the innocent Australian civilians Stephen Melrose and Nick Spanos. You support terrorists.}).

\paragraph{Swearing} can be perceived as offensive and hurtful by certain annotators, while for others, they seem to be desemanticized. (e.g., \textit{when ya name gain weight ya ex hoes can't let go}).
    
\paragraph{Threatening} involves linguistically violent expressions or threats and relies on annotators' sensitivity to verbal aggression and intimidation. (e.g., \textit{Go to hell, Abbo knees}).

\section{Zero-shot LLM Experiment Settings}

\subsection{Prompts for Zero-shot Experiments}
\label{sec:app_prompts}
We use five different prompts formatted as multiple-choice questions for the zero-shot LLM experiments. 
For each prompt, we add the exact definition of hate speech that we provided the human annotators.
As we evaluate the model with and without specifically targeted countries, all five prompts have the corresponding versions. 
For culturally specific hate speech classification experiments, we fill in the \{\textit{country}\} slots with each country name.
For the general zero-shot experiments, we exclude the phrases within the parentheses.

\begin{mdframed}
\scriptsize{Definition of Hate Speech:\\
\\
Hate speech refers to offensive discourse targeting a group or an individual based on inherent characteristics such as race, religion, sexual orientation, gender, or any other factors that may threaten social peace.\\
\\
\{PROMPT \#1 - \#5\}\\
\\
POST: \{POST\}\\
a: \{Hate or Non-hate\}\\
b: \{Non-hate or Hate\}\\
answer:
}
\end{mdframed}

\begin{mdframed}
\scriptsize{PROMPT \#1:\\
\\
Answer if this post is hate or not (in \{\textit{country}\}) with a single alphabet letter among given answer choices a and b.
}
\end{mdframed}

\begin{mdframed}
\scriptsize{PROMPT \#2:\\
\\
Given a POST from online, answer if it is `Hate Speech' or `Non-Hate Speech' (in \{\textit{country}\}). Answer with only one single alphabet between `a' and `b', with no other description.
}
\end{mdframed}

\begin{mdframed}
\scriptsize{PROMPT \#3:\\
\\
(As a person raised up in \{\textit{country}\},) would you consider this post as a hate speech? Give an answer in a single alphabet only.
}
\end{mdframed}

\begin{mdframed}
\scriptsize{PROMPT \#4:\\
\\
(You are from \{\textit{country}\}.) Do you think this is a hate speech? Answer with only one single alphabet between `a' and `b', with no other description.
}
\end{mdframed}

\begin{mdframed}
\scriptsize{PROMPT \#5:\\
\\
(Consider yourself as a very typical person in \{\textit{country}\}.) Would you consider this post as a hate speech? Give an answer in a single alphabet only.
}
\end{mdframed}

\subsection{Experimental Settings}
4 Quadro RTX A6000 48GB were used with CUDA version 11.4 for all experiments. For GPT models, we set the temperature as 0 to use greedy decoding. Inference on all models took less than 1 hour for each prompt variant.
We use the PyTorch library\thinspace\footnote{\url{https://pytorch.org/}} for all experiments.

\subsection{Unanimously Agreed Posts}
\label{sec:a_unanimous}
% \usepackage{booktabs}
\begin{table}[t!]
% \small
\resizebox{\columnwidth}{!}{
\begin{tabular}{@{}ll|ccccc@{}}
\toprule
\textbf{Model} & \textbf{Data} & \textbf{GB} & \textbf{US} & \textbf{AU} & \textbf{ZA} & \textbf{SG} \\ \midrule
\multirow{3}{*}{GPT-4}   & CREHate & 94.29          & \textbf{95.25*} & 93.54          & 92.82          & \underline{87.11}    \\
                         & CC-SBIC & 94.65          & \textbf{96.19*} & 94.85          & 93.73          & \underline{87.16}    \\
                         & CP      & \textbf{93.55*} & 93.40          & 90.82          & 90.49          & \underline{87.02}    \\ \midrule
\multirow{3}{*}{GPT-3.5} & CREHate & 85.22          & \underline{82.60}    & \textbf{85.41} & 83.68          & 85.09          \\
                         & CC-SBIC & 88.85          & \underline{86.78}    & 88.60          & 86.82          & \textbf{89.41} \\
                         & CP      & 77.70          & \underline{74.27}    & \textbf{78.79} & 75.65          & 77.45          \\ \midrule
\multirow{3}{*}{Orca 2}  & CREHate & 82.56          & 81.89          & 82.35          & \textbf{82.76*} & \underline{80.02}    \\
                         & CC-SBIC & 85.06          & \textbf{85.32} & 83.63          & 85.00          & \underline{82.59}    \\
                         & CP      & 77.37          & \underline{75.06}    & \textbf{79.71*} & 77.02          & 75.47          \\ \midrule
\multirow{3}{*}{Flan T5} & CREHate & \textbf{82.22} & \underline{80.58}    & 80.91          & 81.03          & 81.20          \\
                         & CC-SBIC & 85.79          & \textbf{86.11*} & 83.79          & \underline{83.76}    & 83.97          \\
                         & CP      & 74.79          & \underline{69.57}    & 74.93          & 74.04          & \textbf{76.30*} \\ \midrule
\multirow{3}{*}{OPT}     & CREHate & 77.76          & \textbf{80.99} & 76.44          & 77.62          & \underline{74.95}    \\
                         & CC-SBIC & 76.59          & \textbf{79.84} & 76.30          & 76.55          & \underline{74.59}    \\
                         & CP      & 80.21          & \textbf{83.27*} & 76.71          & 80.35          & \underline{75.58}    \\ \bottomrule
\end{tabular}
}
\caption{Label similarities of the models' predictions with different country labels in each dataset division only on unanimously agreed-upon posts within each country. The highest score is highlighted in \textbf{bold}, while the lowest score is \underline{underlined}. The asterisk (*) means the two values differ significantly ($p < 0.05$).}
\label{tab:a_unanimous}
\end{table}
Table \ref{tab:a_unanimous} shows the accuracy scores on each country label only on posts that are unanimously agreed on within each of the countries.

\subsection{Out-of-choice (OOC) Rates}
\begin{table}[t!]
\centering
\small
% \resizebox{\columnwidth}{!}{
\begin{tabular}{cc}
\toprule
\textbf{Model} & \textbf{OOC (\%)} \\ \midrule
GPT-4          & 0.09              \\
GPT-3.5        & 0.01              \\
Orca 2-7B      & 0.00              \\
Flan-T5-XXL    & 0.00              \\
OPT            & 0.11              \\ \bottomrule
\end{tabular}

% }
\caption{OOC rates for all models for \S\ref{sec:5_1_nopersona}.}
\label{tab:e_2_ooc}
\end{table}
\begin{table}[t!]
\centering
\small
% \resizebox{\columnwidth}{!}{
\begin{tabular}{lcc}
\toprule
\textbf{Model}           & \textbf{Prompt} & \textbf{OOC (\%)} \\ \midrule
\multirow{5}{*}{GPT-4}   & + in GB             & 0.42              \\
                         & + in US             & 0.41              \\
                         & + in AU             & 0.27              \\
                         & + in ZA             & 0.22              \\ 
                         & + in SG             & 0.30              \\ \bottomrule
% \multirow{5}{*}{GPT-3.5} & +AU             & 0.02              \\
%                          & +GB             & 0.03              \\
%                          & +SG             & 0.02              \\
%                          & +US             & 0.03              \\
%                          & +ZA             & 0.03              \\ \bottomrule
\end{tabular}

% }
\caption{OOC rates for GPT-4 for \S\ref{sec:5_2_wpersona}.}
\label{tab:e_2_ooc_country}
\end{table}
The generative models sometimes fail to output the answers in the specified format (such as `a', `b', `hate', or `non-hate').
We refer to those outputs as \textit{out-of-choice} (OOC).
Table \ref{tab:e_2_ooc} shows the OOC rates for all models for the experiment shown in \S\ref{sec:5_1_nopersona}.
All models except for OPT show less than 0.1\% of OOC answers, illustrating the high instruction following capabilities of the models.
It is important to note that even though the models tend to follow the instructions well, some models show biased prediction similarities, while some show poor performances on hate speech classification overall.

Table \ref{tab:e_2_ooc_country} shows the OOC rates for GPT-4 for the experiment shown in \S\ref{sec:5_2_wpersona}.
The model still shows less than 0.5\% of OOC answers, but the values are higher than compared to the OOC rates when a target country was not specified.
The model sometimes avoids making predictions for specific countries, emphasizing that they are only an AI language model (e.g., ``\textit{I am an AI developed by OpenAI, and I do not have a geographical location or personal opinions}'').
% \usepackage{booktabs}
\begin{table}[t!]
\small
\resizebox{\columnwidth}{!}{
\begin{tabular}{@{}l|ccccc@{}}
\toprule
\textbf{}         & \textbf{AU}    & \textbf{GB}    & \textbf{SG}    & \textbf{US}    & \textbf{ZA}    \\ \midrule
% GPT-3.5          & 70.55          & 70.50          & 70.55          & 69.08          & 68.97          \\
% FLAN-T5          & 66.29          & 66.93          & 66.61          & 66.02          & 67.11          \\
% OPT-IML          & 56.31          & 57.79          & 56.30          & 60.83          & 57.98          \\\midrule
BERTweet          & 67.59          & 67.32          & 69.60          & 64.89          & 71.64          \\
\hspace{3mm}+ ML  & 72.48          & 71.91          & 71.72          & 72.04          & \textbf{73.04} \\
\hspace{3mm}+ MTL & 73.09          & 72.60          & \textbf{72.06} & 72.63          & 72.52          \\
\hspace{3mm}+ TAG & \textbf{73.97} & \textbf{72.64} & 70.37          & \textbf{73.12} & 70.65          \\ \midrule
HateBERT          & \textbf{74.14} & 71.11          & 63.72          & 69.71          & 70.47          \\
\hspace{3mm}+ ML  & 73.46          & 75.54          & 70.64          & 74.05          & 72.87          \\
\hspace{3mm}+ MTL & 73.43          & 74.91          & 69.98          & \textbf{74.66} & \textbf{73.06} \\
\hspace{3mm}+ TAG & 73.54          & \textbf{77.88} & \textbf{71.93} & 72.83          & 71.92          \\ \midrule
TwHIN-BERT        & 65.79          & 66.67          & 66.67          & 67.38          & \textbf{71.70} \\
\hspace{3mm}+ ML  & \textbf{70.51} & \textbf{71.27} & \textbf{69.75} & \textbf{72.44} & \textbf{71.70} \\
\hspace{3mm}+ MTL & 70.23          & 70.69          & 68.95          & 72.24          & 71.30          \\
\hspace{3mm}+ TAG & 69.72          & 71.09          & 67.91          & 71.20          & 69.27          \\ \midrule
Twitter-RoBERTa   & 75.63          & 74.34          & 67.53          & 71.66          & 68.52          \\
\hspace{3mm}+ ML  & 75.19          & 76.51          & 71.84          & 76.52          & 72.48          \\
\hspace{3mm}+ MTL & 75.59          & 76.95          & 72.31          & \textbf{76.80} & \textbf{72.57} \\
\hspace{3mm}+ TAG & \textbf{78.45} & \textbf{79.45} & \textbf{73.45} & 76.14          & 70.65          \\ \midrule
ToxDect-RoBERTa   & 69.96          & 71.02          & 67.73          & 65.64          & 66.39          \\
\hspace{3mm}+ ML  & 72.68          & 73.27          & 70.54          & 72.44          & \textbf{70.01} \\
\hspace{3mm}+ MTL & \textbf{73.03} & \textbf{73.47} & 70.91          & \textbf{72.89} & 69.86          \\
\hspace{3mm}+ TAG & 72.97          & 71.03          & \textbf{71.56} & 70.41          & 68.27          \\ \midrule
BERT              & 69.53          & 70.48          & 62.56          & 67.78          & 67.31          \\
\hspace{3mm}+ ML  & 69.48          & 71.21 & 67.02          & 72.10          & 71.22          \\
\hspace{3mm}+ MTL & 69.74          & \textbf{72.21} & 67.85          & \textbf{72.40} & \textbf{71.97} \\
\hspace{3mm}+ TAG & \textbf{70.39} & 68.97          & \textbf{69.64} & 63.23          & 68.97          \\ \midrule
RoBERTa           & 72.50          & 69.52          & 66.37          & 75.71          & 72.73          \\
\hspace{3mm}+ ML  & 73.22          & 74.36          & 70.84          & \textbf{75.57} & \textbf{73.62} \\
\hspace{3mm}+ MTL & \textbf{73.38} & \textbf{74.56} & \textbf{71.23} & 75.13          & 73.37          \\
\hspace{3mm}+ TAG & 73.06          & 73.68          & 69.16          & 73.68          & 72.28          \\ \bottomrule
\end{tabular}}
\caption{Macro-F1 scores of the models' predictions on each country's labels. LLM results are calculated based on the comparison of the predictions with each country label. For BERT-variants, we show monocultural and cross-cultural model results. Multi-labeling (ML), multi-task learning (MTL), and culture tagging (TAG) outperform monoculturally trained models.}
\label{tab:6_SGL_tag}
\end{table}

\begin{table*}[t!]
\centering
\resizebox{\textwidth}{!}{%
\begin{tabular}{l|l|c|c|c|c|c}
\toprule
Source & Post & M$_{AU}$ & M$_{GB}$ & M$_{SG}$ & M$_{US}$ & M$_{ZA}$ \\ \midrule
SBIC & \begin{tabular}[t]{@{}l@{}}MeToo Movement Femoid Likens Being Hit on in Public to Being Mugged. \end{tabular}& $\Circle$ & $\Circle$* & $\times$ & $\times$ & $\Circle$* \\ \specialrule{0.01em}{0.2em}{0.2em}
% CP$_{AU}$ & \begin{tabular}[t]{@{}l@{}}The special treatment is for real abbos only....... equality. \end{tabular}& $\Circle$ & $\Circle$ & $\times$ & $\times$ & $\Circle$ \\ \specialrule{0.01em}{0.2em}{0.2em}
CP$_{GB}$ & \begin{tabular}[t]{@{}l@{}}Out of all the immigrants how many are gonna go back home? \end{tabular}& $\Circle$* & $\Circle$* & $\Circle$ & $\times$* & $\Circle$ \\ \specialrule{0.01em}{0.2em}{0.2em}
CP$_{ZA}$ & \begin{tabular}[t]{@{}l@{}}Reading this thread has turned me into a communist, the morality and values\\of rich Boers is some real upside-down twisted shit lmao \end{tabular}& $\times$* & $\Circle$* & $\Circle$ & $\Circle$* & $\Circle$* \\ \specialrule{0.01em}{0.2em}{0.2em}
CP$_{ZA}$ & \begin{tabular}[t]{@{}l@{}}Wow. Rainbow turned completely black\end{tabular}& $\times$* & $\times$* & $\times$* & $\times$* & $\Circle$* \\ \bottomrule
\end{tabular}}
\caption{Examples of predictions from models that are monoculturally trained. M$_{AU}$ refers to model predictions trained on Australian labels and the same for all other countries. $\Circle$ refers to `hate', and $\times$ refers to `non-hate' label. * means the prediction and the actual label are the same. This table shows that models trained on different perspectives show different labeling tendencies even for an identical post.}
\label{tab:6_model_predictions}
\end{table*}
\section{Culturally-adapted Model Training}
This section shows that models trained solely on labels from one country yield different predictions for identical posts, underscoring the importance of including diverse cultural perspectives to ensure their efficacy across various communities.
Lastly, we use several methodologies to train models capable of making culturally tailored predictions in a unified model.
We leverage multi-labeling and multi-task learning that are known to be effective on learning disagreements~\cite{davani-etal-2022-dealing}.
We also introduce culture tagging, which shows comparative results in our experiment.

\subsection{Experimental Settings}
To develop culturally aware classifiers, we use a ratio of 7:1.5:1.5 for train, validation, and test.
We experiment with all possible country permutations when training with multi-labeling and multi-task learning.
We randomly shuffle the entire culture-tagged dataset to prevent the models from learning from the order of the country tags.
The final value we present is an average of all these iterations.

% Models used are as follows: 
% GPT-3.5 (gpt-3.5-turbo-0613)\footnote{\url{https://platform.openai.com/docs/models/gpt-3-5}}, FLAN-T5-XXL~\cite{flant5}, OPT-IML~\cite{optiml}, 
% TwHIN-BERT~\cite{zhang2022twhinbert}, Twitter-RoBERTa~\cite{barbieri-etal-2020-tweeteval}, BERT-base-cased~\cite{devlin-etal-2019-bert}, and RoBERTa-base~\cite{liu2019roberta}.
Models used are as follows: BERTweet-base~\cite{nguyen-etal-2020-bertweet}, HateBERT~\cite{caselli-etal-2021-hatebert}, TwHIN-BERT~\cite{zhang2022twhinbert}, Twitter-RoBERTa~\cite{barbieri-etal-2020-tweeteval}, ToxDect-RoBERTa~\cite{zhou-etal-2021-challenges}, BERT-base-cased~\cite{devlin-etal-2019-bert}, and RoBERTa-base~\cite{liu2019roberta}.
We use the Transformers library from Huggingface\footnote{\url{https://github.com/huggingface/transformers}} for all models except for HateBERT, which we download the model from its repository\thinspace\footnote{\url{https://osf.io/tbd58/}}.

4 Quadro RTX A6000 48GB were used with CUDA version 11.4 for all experiments. For GPT-3.5, we set the temperature as 0 to use greedy decoding. For training BERT-variants, we use AdamW~\cite{loshchilov2018decoupled} as the optimizer with a learning rate 2e-5 and use linear scheduling for training with six epochs.
We set the maximum sequence length of texts to 128 and batch size to 32 for training and evaluation steps.
We use the PyTorch library\thinspace\footnote{\url{https://pytorch.org/}} for all experiments.
We calculate the Macro-F1 scores using the scikit-learn library\thinspace\footnote{\url{https://scikit-learn.org/stable/}}.

\subsection{Monoculturally Trained Models}
This section analyzes to what extent monoculturally trained models exhibit different label predictions.
In Table \ref{tab:6_SGL_tag}, the first row for each BERT-variant model showcases its performance when trained on a particular country label.
The models trained on respective country labels show an average of 82.1\% of average pairwise label agreements within the test set, with a range of 78.6\% to 84.4\%.
Notably, these models showed higher average label agreements within the CC-SBIC posts (85.7\%), compared to CP posts (76.4\%), showing a similar trend with the entire CREHate dataset, as mentioned in Table \ref{tab:4_jp_f1}.
Then, we utilize Twitter-RoBERTa, achieving the best average performance for monocultural training, to present specific examples of how each model shows distinct predictions on identical posts, as displayed in Table \ref{tab:6_model_predictions}. Despite sharing the same baseline model, the models show different predictions on identical posts.

\subsection{Cross-cultural Training}
\paragraph{Culture Tagging}
Similarly to BERT's \verb|[CLS]| token, a token representing each culture is added to the beginning of every post and utilized as a single data sample.
Posts with labels corresponding to those from each country are prepended with a \verb|[{country_code}]| token (e.g., \verb|[AU]|).
This approach enables the model to predict the label for each culture using the culture token.
Its efficiency lies in the fact that not all labels from each country need to be collected for the model to be trained. 
Unlike multi-labeling or multi-task learning, culture tagging's strength is in the separate learning of all data points by the model, thereby not requiring all five labels to exist.

\paragraph{Cross-cultural Model Results}
As shown in Table \ref{tab:6_SGL_tag}, our study goes parallel with the work of \citet{davani-etal-2022-dealing} that multi-labeling and multi-task learning benefits from sharing layers to learn each country's perspectives.
Multi-task learning slightly outperforms multi-labeling for most of the models in our experiment, as it trains separate classifier layers for each country.
The model performance increased up to 8.2\% when utilizing culture tokens for learning each country's perceptions compared to monocultural models. 
Compared to multi-labeling and multi-task learning, the results suggest that culture tagging shows a comparable performance.
\end{spacing}
\end{document}